# Crossing Roads of Federated Learning and Smart Grids: Overview, Challenges, and Perspectives

Hafsa Bousbiat[a], Roumaysa Bousselidj[b], Yassine Himeur[c,*], Abbes Amira[d,e], Faycal Bensaali[f], Fodil Fadli[g], Wathiq Mansoor[c] and Wilfried Elmenreich[h]

[a]*University of Applied Sciences, Institut für Informatic, Wiener Neustadt, 2700, Austria*
[b]*University of Skikda, Department of Computer Science, Mezghich, Skikda, 21000, Algeria*
[c]*University of Dubai, College of Engineering and Information Technology, Dubai, United Arab Emirates*
[d]*University of Sharjah, College of Computing and Informatics, Sharjah, United Arab Emirates*
[e]*De Montfort University, Leicester, Institute of Artificial Intelligence, United Kingdom*
[f]*Qatar University, Department of Electrical Engineering, Doha, 2713, Qatar*
[g]*Qatar University, College of Engineering, Department of Architecture and Urban Planning, Doha 2713, Qatar*
[h]*University of Klagenfurt, Smart Grids Research Group, Klagenfurt 9020, Austria*

ABSTRACT

Consumer's privacy is a main concern in Smart Grids (SGs) due to the sensitivity of energy data, particularly when used to train machine learning models for different services. These data-driven models often require huge amounts of data to achieve acceptable performance leading in most cases to risks of privacy leakage. By pushing the training to the edge, Federated Learning (FL) offers a good compromise between privacy preservation and the predictive performance of these models. The current paper presents an overview of FL applications in SGs while discussing their advantages and drawbacks, mainly in load forecasting, electric vehicles, fault diagnoses, load disaggregation and renewable energies. In addition, an analysis of main design trends and possible taxonomies is provided considering data partitioning, the communication topology, and security mechanisms. Towards the end, an overview of main challenges facing this technology and potential future directions is presented.

## Abbreviations

| | |
|---|---|
| **AI** | Artificial Intelligence |
| **AMI** | Advanced Metering Infrastructure |
| **CS** | Charging Station |
| **EDO** | Energy Data Owner |
| **ESP** | Energy Service Provider |
| **EV** | Electrical Vehicle |
| **FDI** | False Data Injection |
| **FL** | Federated Learning |
| **HFL** | Horizontal Federated Learning |
| **IID** | Independently Identically Distributed |
| **LR** | Linear Regression |
| **ML** | Machine Learning |
| **MPC** | Multiparty Computation |
| **NILM** | Non-Intrusive Load Monitoring |
| **P2P** | Peer-To-Peer |
| **PV** | Photo-Voltaic |
| **RE** | Renewable Energy |
| **RL** | Reinforcement Learning |
| **SG** | Smart Grid |
| **S** | Supervised |
| **TFL** | Transfer Federated Learning |
| **V2G** | Vehicle-to-Grid |
| **V2V** | Vehicle-to-Vehicle |
| **VFL** | Vertical Federated Learning |
| **VPP** | Virtual Power Plant |

*Corresponding author

hafsa.bousbiat@gmail.com (H. Bousbiat); er_bousselidj@esi.dz (R. Bousselidj); yhimeur@ud.ac.ae (Y. Himeur); aamira@sharjah.ac.ae (A. Amira); f.bensaali@qu.edu.qa (F. Bensaali); f.fadli@qu.edu.qa (F. Fadli); wmansoor@ud.ac.ae (W. Mansoor); wilfried.elmenreich@aau.at (W. Elmenreich)

ORCID(s): 0000-0003-0338-732X (H. Bousbiat); 0000-0003-1324-3346 (R. Bousselidj); 0000-0001-8904-5587 (Y. Himeur); 0000-0003-1652-0492 (A. Amira); 0000-0002-9273-4735 (F. Bensaali); 0000-0002-7917-0939 (F. Fadli); 0000-0003-2784-5188 (W. Mansoor); 0000-0001-6401-2658 (W. Elmenreich)

## 1. Introduction

Smart Grid (SG) is a revolutionary concept introduced to join power systems' production, distribution, and generation in a single frame. The ultimate goal of SG consists of the continuous and efficient delivery of energy to the end-users through advanced communication and sensing technologies combined with Artificial Intelligence (AI) algorithms for nearly real-time monitoring of power generation and consumption [1]. Consequently, an increased number of edge devices augmented with intelligent decision-making functionalities are deployed [2, 3], generating a huge amount of data and leading to new challenges such as computational resources, communication bandwidth, security, and privacy concerns [4].

A viable solution to address the previous issues consists of edge-enabled AI technologies, notably Federated Learning (FL) [5, 6, 7]. FL was first proposed by Google in 2017 as a new form of privacy-preserving distributed learning [8]. FL helps protect sensitive data generated by SG components by training Machine Learning (ML) models on local data rather than transmitting it to a centralized server [9]. This can reduce the risk of data breaches and ensure that sensitive information is not exposed to unauthorized parties. SG components generate vast amounts of data that can be used to improve the performance and efficiency of the system [10]. It also helps increase the system's scalability by distributing the computation workload across multiple nodes, reducing the need for centralized processing [11]. Moreover, by training ML models on local data, FL reduces the amount of data to transmit over the network and hence improves the efficiency of the SG by reducing network congestion and improving response times [12]. Additionally, by training models on data from multiple sources, FL can help create models more representative of the entire system and better adapt to changing conditions [13].

In pursuit of this goal, significant emphasis has been placed on SG frameworks based on FL. Recently, a growing number of research articles have highlighted the crucial need for a review article that can enhance the research topic by accomplishing the following tasks: (i) identifying gaps in FL-based SG research; (ii) evaluating the quality and validity of research studies and recognizing their strengths and limitations; (iii) critiquing methodologies, providing analysis, and proposing alternative or improved methods; (iv) facilitating the improvement of research design by offering constructive feedback and suggestions; and (v) sharing research findings, and deriving recommendations.

### 1.1. Paper's Contributions

To date, several studies separately investigating either SG services or FL already exist in the literature [14, 15, 16]. Nonetheless, no review article has been found that discusses the use of FL in SG. Typically, very little is known about existing and state-of-the-art frameworks coupling the two scholarships, with no detailed analysis of the advantages and limitations of such technology. The current manuscript addresses this gap by closely examining the applications of FL in SG setups. The main contributions of the current paper are summarised as follows:

- A presentation of FL fundamental concepts including: (i) the definition of FL, (ii) categories of FL (iii) communication topologies of FL, and (iv) evaluation of FL.
- A discussion of the degrees of freedom and research trends of FL-based frameworks for SG, analyzing their advantages and drawbacks.
- A comprehensive literature review of existing FL-based frameworks for the different services related to the management of SG with a comparative analysis of the achieved results for each service.
- A discussion of implementation challenges by summarizing the lessons learned, critical open research questions, and future research direction.

### 1.2. Methodology

The literature search considered four search databases: Scopus, IEEE, ACM library, and Google Scholar with the following query: *"federated learning"* AND *("energy"* OR *"smart grid"* OR *"energy forecasting"* OR *"load disaggregation"* OR *"thermal comfort control")*. The search methodology resulted in 286 contributions refined to 120 at the end, as explained in Figure 1. The duplicates were first eliminated, followed by a screening process based on the title, the abstract, and, later on, based on the full text. A final set of 120 contributions were considered. Figure 1 further illustrates the resulting word cloud from the analysis of the keywords. It reveals three main topics: *Energy applications*, *Security and privacy*, and the *Modeling approach*. Energy applications include load forecasting, demand response, energy efficiency, renewable energy, load disaggregation, and smart contracts. Regarding security, blockchain is one of the most frequently used keywords, along with data privacy, security, and poisoning attacks. The third topic includes keywords such as AI, machine learning, LSTM, deep neural networks, and convolutional neural network. The remainder of the paper discusses related work based on an extended set of these topics.

### 1.3. Organization

The remainder of the current manuscript proceeds as follows: Section 2 introduces basic concepts related to FL. Section 3 describes the main design dilemmas and research trends of the application of FL in SG. Section 4 discusses existing FL-based frameworks for the energy services in SG. Section 5 highlights this technology's primary challenges with possible future directions to address them discused in Section 6. Finally, a conclusion of the current study is presented in Section 7.

## 2. Background

### 2.1. FL Paradigm

FL is an ML strategy that relies on training algorithms across various decentralized edge servers or devices holding local data patterns without the need to exchange them. This technique is applied in contrast to traditional centralized ML schemes, in which all local data repositories are uploaded to a central server. Accordingly, FL allows various actors to build a robust ML algorithm without sharing data to overcome crucial challenges such as privacy preservation, right of access to sensitive data, and access to heterogeneous data. For this purpose, FL involves the following four main learning steps [17, 18]:

a) **Global model initialisation**: The global model is randomly initialized at the beginning of the training process.
b) **Client selection**: Due to constraints on the communication network, only a subset of randomly selected nodes contribute to the training at each round. In some cases, the selection further considers network reliability (e.g., [19]) to avoid computational delay.
c) **Local training**: Each selected client $i$ trains the model on its local dataset $D_i$ for a certain number of epochs $N$. The updated model is then uploaded to the central node in the case of a centralized topology or broadcast to all the other nodes in the case of a decentralized topology.
d) **Model aggregation**: At the reception of locally trained models, an updated global model is built considering a weighted average of the received models where the weights represent the Amount of data of each client $i$. The new global model is sent back to the different nodes.

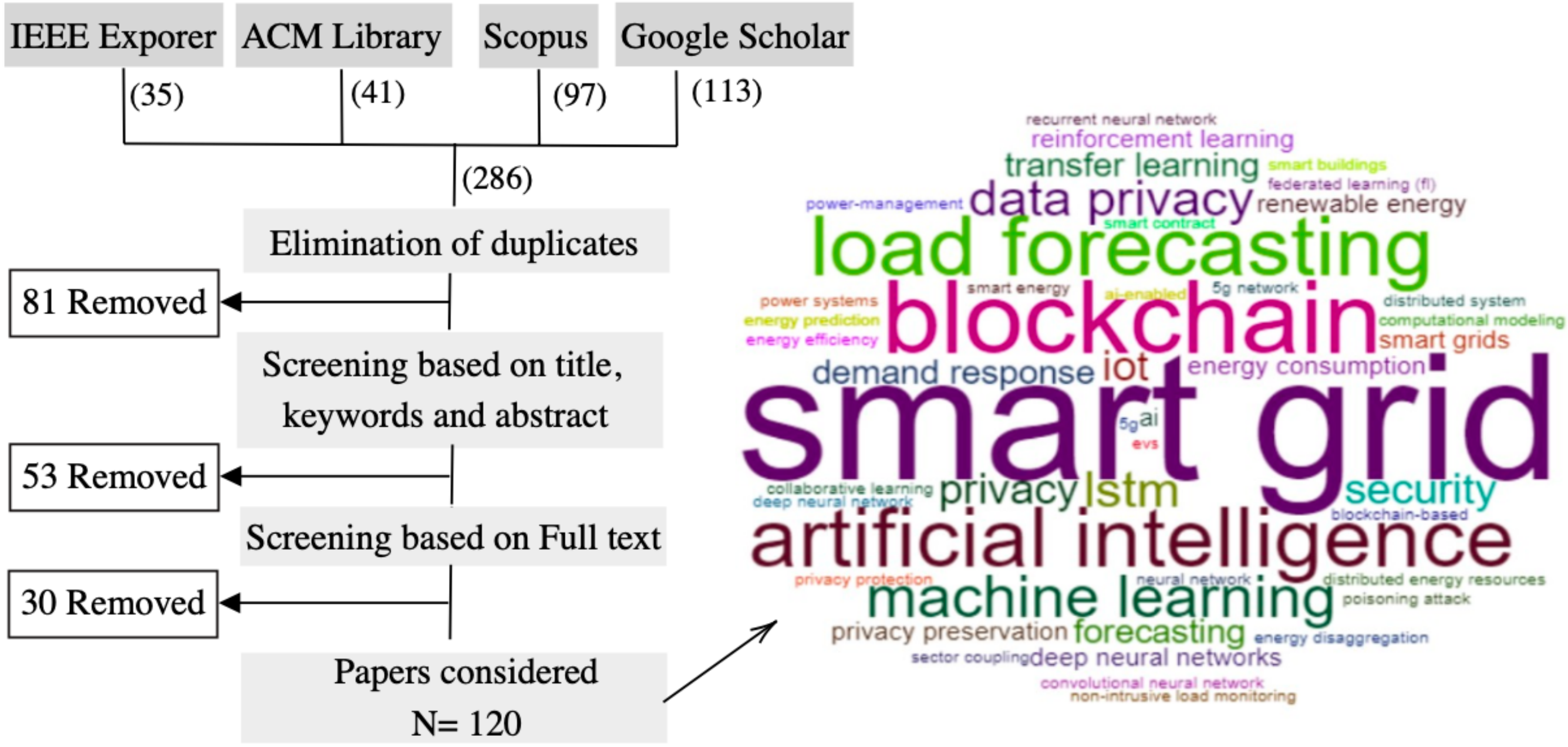

**Figure 1:** Literature review process and keywords cloud analysis

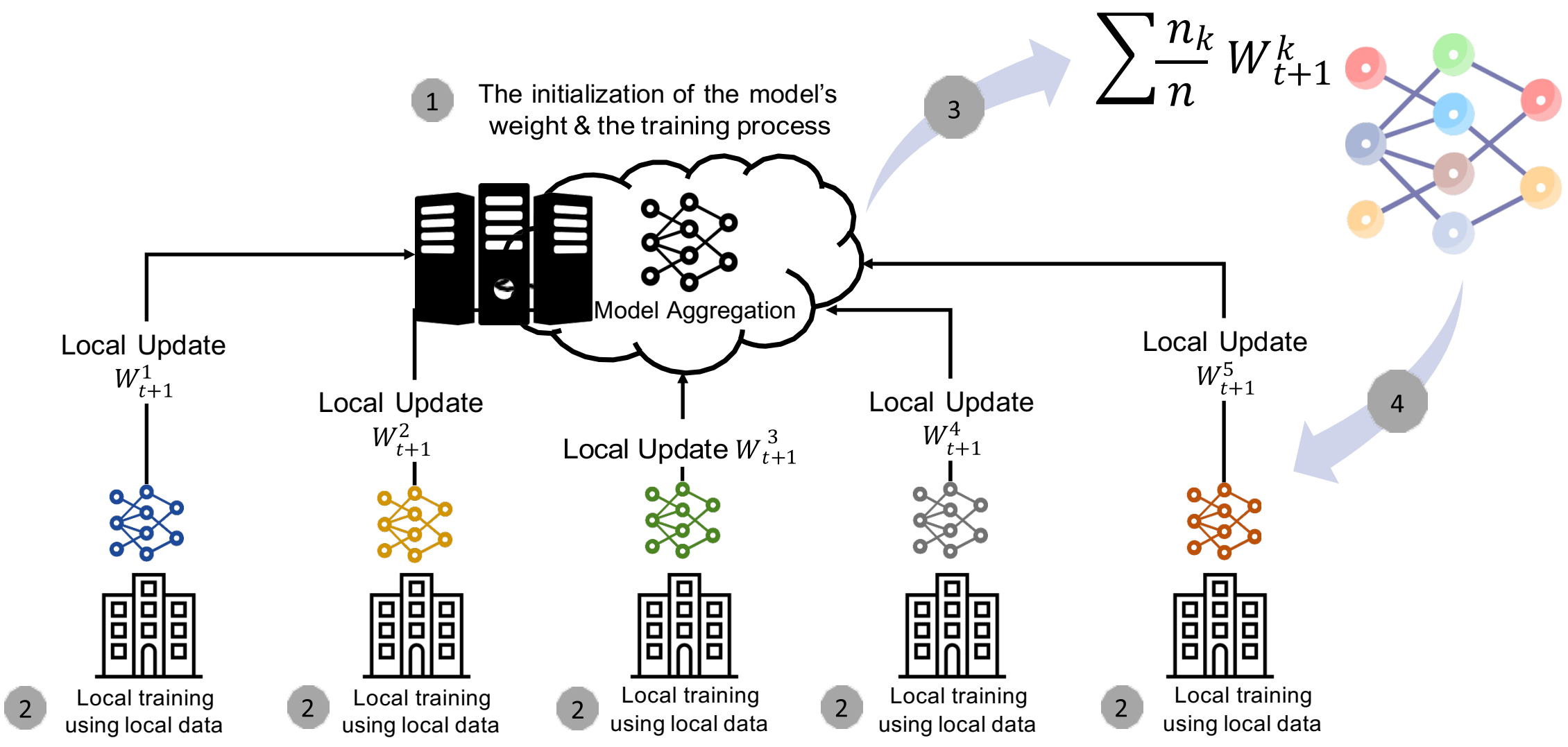

**Figure 2:** Overview of learning steps in FL (*image source [23]*)

The steps *b* to *d* are repeated until convergence or until a maximum number of rounds is achieved. The previously described aggregation algorithm is commonly referred to as FedAvg [20] algorithm. Unlike centrally trained models, evaluating FL frameworks requires examining additional aspects due to the distributed nature of the training process that is sensitive to several attacks. Several evaluation frameworks exist in the literature [21, 22]. Considering the characteristics of SG, we provide an extended list of these evaluation aspects in Table 1. Figure 2 summarizes the learning steps adopted in FL frameworks.

### 2.2. Categories of FL

Three different categories of FL are identified [18, 24, 25, 26, 27] based on the data partitioning approach: (i) horizontal FL, (ii) vertical FL, and (iii) transfer FL. For clarity purposes, this section briefly discusses the three categories. ***Horizontal Federated Learning (HFL)***, also referred to as sample-based FL [24], is used in scenarios where local datasets $D_i$ have the same feature space but different ID space [18, 26]. More precisely, HFL is applicable in cases where the local datasets $D_i$ have the same structure but for different data samples. ***Vertical Federated Learning (VFL)***, also referred to as feature-based FL [24], is applicable in scenarios where the local datasets $D_i$ have different feature space with an overlapping ID space [18, 26]. As such, VFL is applicable in cases where the local datasets $D_i$ have partial features and an overlapping set of data samples. In contrast with HFL, VFL can turn out to be more complicated as it requires entity alignment [27]. The latter consists of finding records from two databases representing the same real-object [28]. Figure 3 illustrates the previous two approaches. ***Transfer Federated Learning (TFL)*** is applicable in scenarios where local datasets $D_i$ have different feature space and/or different ID space [18, 26]. It can further be sub-categorized into: (1) Instance-based TFL, (2) Feature-based TFL, and (3) Parameter-based TFL.

**Table 1**
Evaluation of FL frameworks

| Aspect | Description | Measures |
|---|---|---|
| **Convergence** | Evaluates the required resources to achieve acceptable performance | Number of communication rounds, training time |
| **Communication efficiency** | Evaluates the amount of data that needs to be transmitted between devices during the communication rounds | Amount of data transmitted, transmission time |
| **Data privacy and security** | Ensures that sensitive data is not compromised or leaked | Testing the model against possible attacks |
| **Model accuracy** | Evaluates the model's performance on different scenarios | Measures the model's performance on both Independently Identically Distributed (IID) and non-IID data |
| **Robustness** | Evaluate the system's performance under different parameters and conditions | Ablation and hyper-parameters studies |

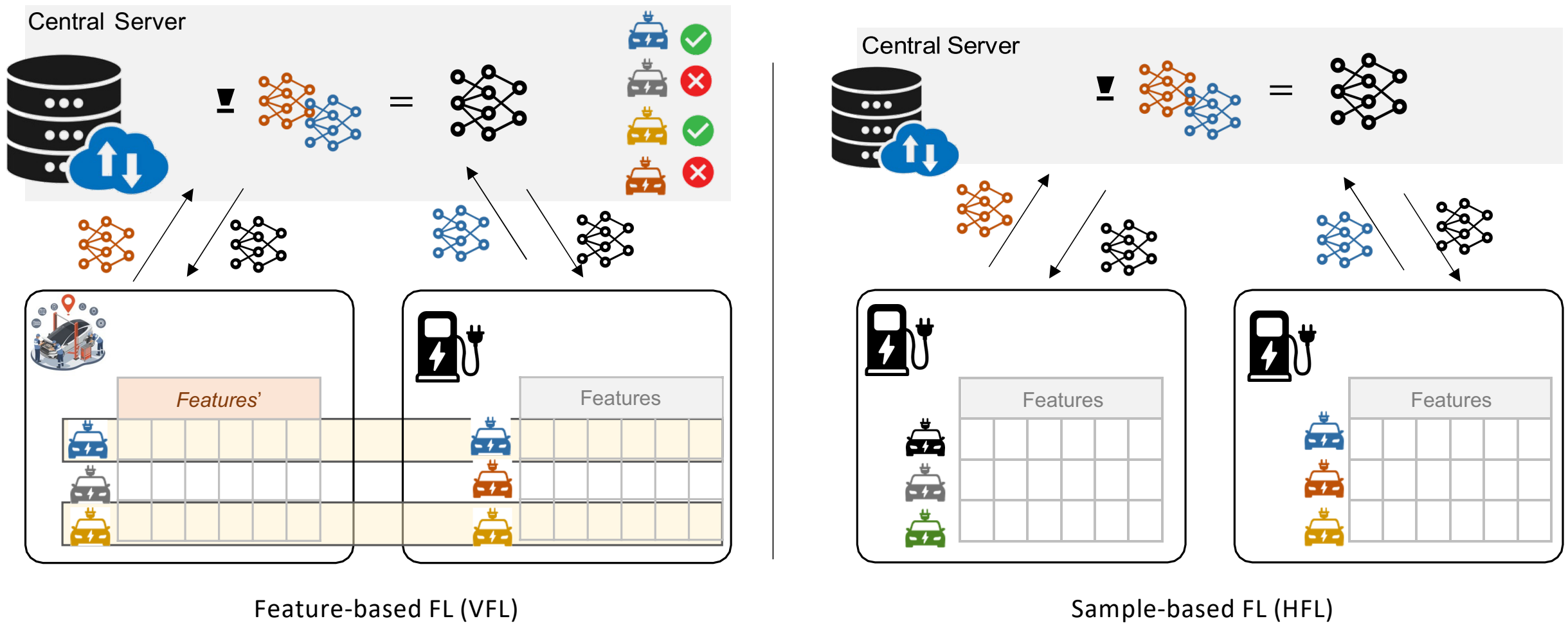

**Figure 3:** Data partitioning in FL for EV and CS

## 2.3. Topologies of FL

The communication architecture between distant clients in FL is a crucial aspect influencing performance. Considering related work, three main communication topologies can be identified: centralized FL, decentralized FL, and hierarchical FL. Figure 4 compares the first two topologies and classical learning. The hierarchical topology is not represented as it could be considered a variant of centralized learning with an extra communication hop. ***Centralised FL***, also referred to as client-server architecture, corresponds to a scenario where $N$ data owners (i.e., clients) locally train the same models using different local datasets $D_i$ with a central node aggregating these models. This architecture assumes that the clients are honest participants and only the server is honest, but curious [24]. ***Decentralised FL***, also referred to as Peer-to-Peer (P2P) architecture, corresponds to a scenario where $N$ data owners train the same model using different local datasets $D_i$ with direct communication between these nodes to construct an updated model after each iteration. P2P topology is thus a serverless topology where the learning does not require a central node to orchestrate the learning process [29, 30]. Furthermore, the FL paradigm is referred to as *Gossip learning* [31] when this architecture is adopted.

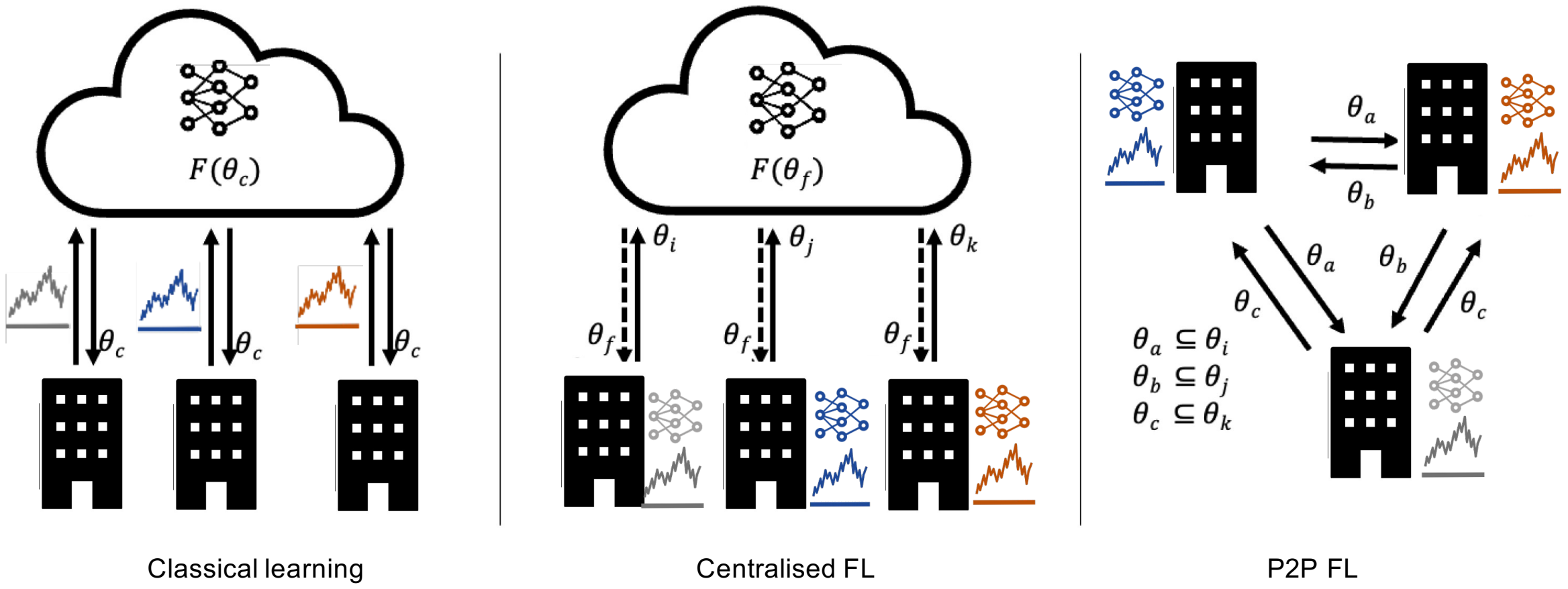

**Figure 4:** Communication Topologies of FL

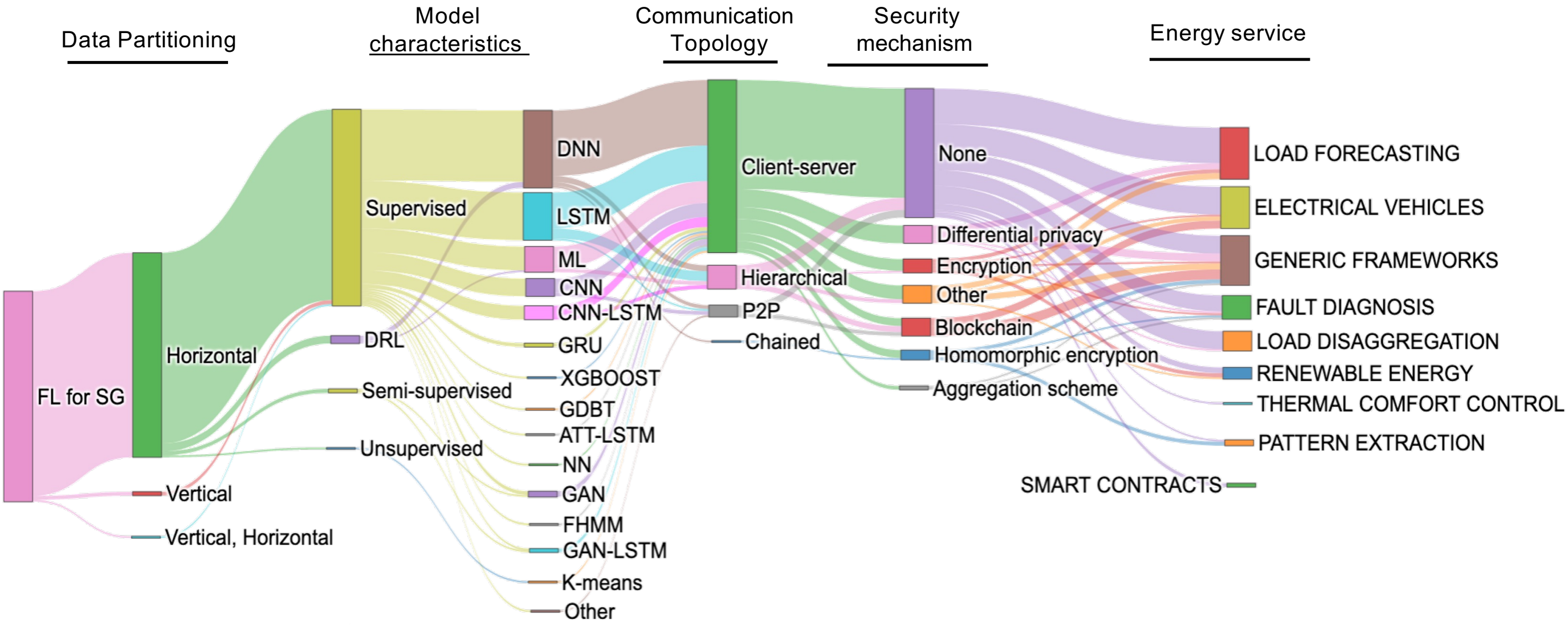

**Figure 5:** Research trends and taxonomies of FL contributions in SGs

## 3. Overview of FL frameworks in SGs

This section describes the main research trends and characteristics of FL frameworks in SGs. Five aspects are considered, including three that already emerged from the word cloud in Figure 1, that is *energy services*, *security and privacy*, and *modeling approach*. Two additional aspects are further investigated, related to the FL paradigm, that is, the *data partitioning*, and the *network topology*. Figure 5 illustrates key variants that emerged during the review process for each considered aspect and the relationship between them. The first four aspects are discussed in the following subsections. However, the application of FL for different energy services is presented in Section 4.

### 3.1. Data Partitioning

The majority of the reviewed contributions show strong evidence for HFL. More precisely, 98% of the reviewed FL-frameworks for SG adopt horizontal partitioning. Considering the characteristics of SG, this trend can be justified with the ownership of energy data by a single Energy service provider (ESP). HFL is often combined with transfer learning. One principal subcategory is mainly adopted, the parameter-based, while little consideration of instance-based and feature-based TFL. Parameter-based TFL consists of: (1) fine-tuning a global model after each training round on the level of the clients, also referred to as *personalization*, or (2) initializing the global model with a model pre-trained on another domain [**?** ]., The first variant was primarily adopted in several works [32, 33] to address the data's non-IIDness and enhance the performance of local models. Despite the success of the previous technique, little attention was dedicated to studying the computational delay that this technique may induce on the computational time, considering the heterogeneity of hardware characteristics of edge devices. The second variant was only adopted in reference [34], where the experimental setup demonstrated a strong need for fine-tuning.

An exception to the previous trend can be observed in the case power prediction of EV, and CS as suggested per references [35, 36] where VFL was prominent. The primary purpose of adopting the VFL is to obtain a complete record of each car driver through the combination of features recorded in the EV system and features recorded at the CS level. The development of secured VFL is still in its infancy, and the majority of security mechanisms applied are only valid with simple ML models [27], basically linear models [18]. Consequently, this scheme is rarely used in SGs. In references [35, 36], it was combined with cryptographic alignment to mitigate the previous issue. However, it is expected that this counter-measure would increase the computational complexity.

### 3.2. Model characteristics

The *supervised learning* paradigm was the most commonly used among the reviewed contributions, with eighty-one (81) contributions in total, that is 72%, adopting this paradigm, mainly with deep neural networks for energy modeling. While adopting this learning paradigm is viable in the case of some energy services, such as load forecasting and power production generation, the availability of labeled data remains a huge obstacle in the case of energy services, such as anomaly detection and load disaggregation. Even if applicable in laboratory and simulation setups, the second case is merely impossible in real smart grid setups. A viable alternative, in this case, is the adoption of semi-supervised algorithms. It was examined in [37], showing competitive results for fault detection. Nonetheless, the experimental setup included only tests on a simulated dataset. Thus, further experiments are required before a clear conclusion can be established about the effect of this paradigm on the performance.

In contrast, *reinforcement and unsupervised learning* received less attention. Reinforcement Learning (RL) was adopted in four (04) [34, 38, 39, 40] different contributions for four (03) different services, that is low voltage control in distribution networks, demand response in distribution networks, and energy management in EV networks. All these contributions demonstrated comparable results to supervised approaches. The only contribution suggesting an unsupervised learning approach leveraging the k-means algorithm was suggested in reference [41] for the case of extracting power consumption patterns from load profiles.

### 3.3. Communication Topology

A closer inspection of energy-related work reveals different and heterogeneous topologies in SG derived from the two most popular FL topologies presented earlier. Figure 6 illustrates the main topologies identified in the studies included in our literature search.

The client-server topology was the most commonly used in SG setups with its two variants: (1) standard client-server and (2) clustered client-server. Studies have adopted the first topology as a typical architecture. However, it could lead to convergence issues due to the heterogeneous load profiles exhibited by different clients [42, 43, 44]. The clustered variant was thus suggested to address this limitation by using several central nodes, optionally communicating with each other, each responsible for aggregating only a subset of locally trained models to mitigate the effect of non-IIDness of the energy data.

Only a few FL studies adopt a peer-to-peer topology to overcome the single failure point of other alternatives by allowing each node to act as a client and a server simultaneously. For example, this architecture was adopted in [45] where the authors consider clients to be prosumers (i.e., a producer and consumer of energy, for example a house with a PV system) to extend energy trading platforms with the concept of Peer-To-Peer (P2P) energy sharing. This topology is particularly interesting for small energy communities. Nonetheless, it depends on the hardware resources available at the edge since the aggregation is performed locally. Furthermore, they require a high communication bandwidth since each client broadcasts its local model to the rest of the clients.

A major challenge for the previous two topologies is the non-IIDness of energy data due to its dependency on the consumer's behavior [46], particularly in the residential sector. To deal with this problem, the edge computing layer can be leveraged. The hierarchical topology was thus adopted in several frameworks. This topology introduces an extra communication hop in the standard client-server topology, where an intermediate aggregation is performed. The reason behind this aggregation is to group consumers according to the similarity of their energy consumption data [42], which allows for mitigating the convergence issues. Despite adopting this topology, a consensus on the best clustering techniques/criteria is still not established. The conducted review revealed the following criteria:

- *Localisation* is the most popular criterion. It was adopted in eight (8) frameworks out of twelve (12) total energy-related hierarchical FL frameworks. The intuition behind adopting longitude and latitude as clustering criteria is that buildings/clients in the same close regions will have the same weather conditions, resulting in similar usage patterns and, thus, similar energy consumption.
- *Attributes*
    1. *Attributes of the clients* also consist of viable clustering criteria where the intuition behind it is that clients with different socio-economic features (e.g., employee/retired or family/students) are likely to exhibit similar load profiles [47] leading to faster convergence.

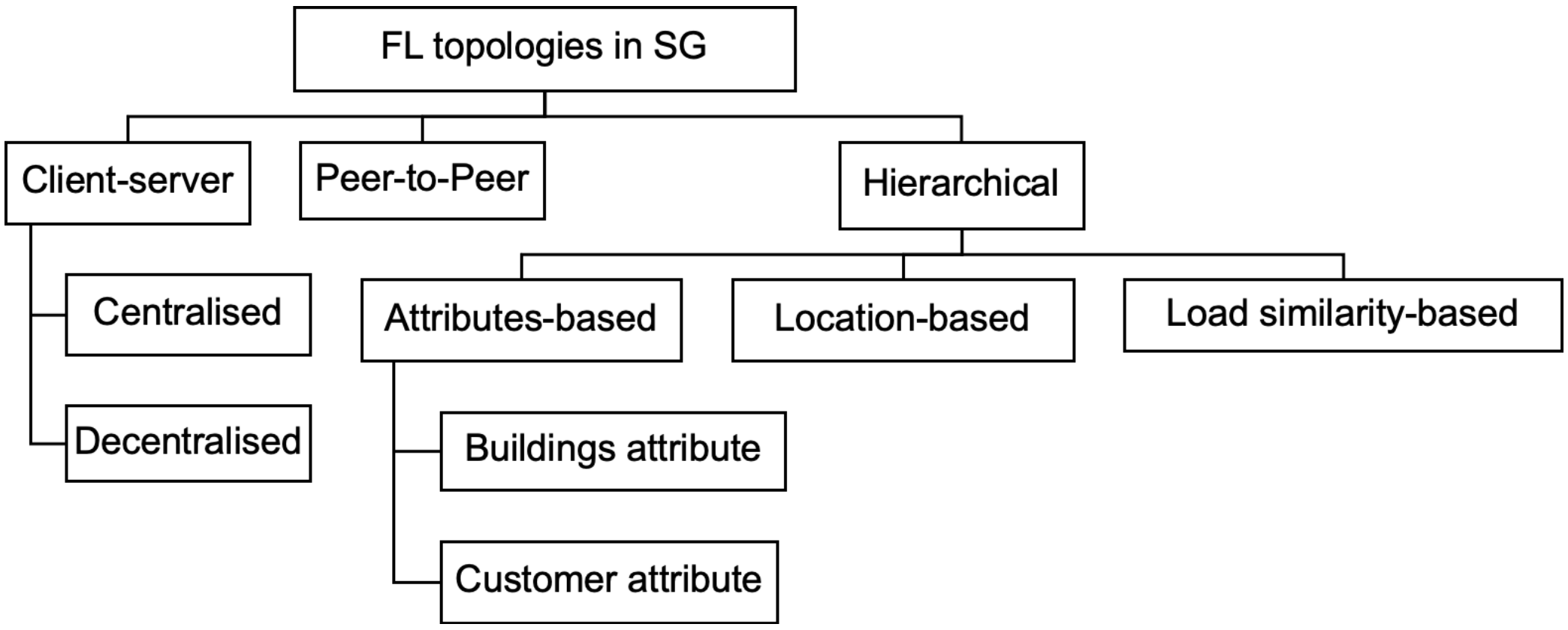

**Figure 6:** Communication topologies in SG-FL frameworks

2. *Attributes of the building*: were adopted in reference [46] including building type, facing direction, region, rental units, and heating type. Building attributes define the consumer's interaction with major loads composing the power load, such as the heating system.

- *Load profiles* was suggested in reference [47], where the authors argued that the direct use of the power curve allows for overcoming convergence issues and is more efficient compared to other clustering criteria. More precisely, the authors suggest using low-frequency energy data collected by the ESP for billing purposes to perform clustering while using high-frequency data locally stored for the training.

- *Parameters of local models* leveraging unsupervised clustering was adopted in reference [42] where the clustering is based on the hyper-parameters of local neural networks trained on edge devices. Although the proposition showed high potential in accurately clustering the clients, it is limited by its high dependency on the capabilities of the client devices for performing the local training.

- *Membership value* allows evaluating the membership of each client to each cluster considering different training rounds. This approach results in dynamic clustering at each training round considering the potential change in consumption, particularly applied in the case of predicting the energy consumption of charging stations [48].

All the previously defined topologies have a direct influence on the aggregation algorithm and, thus, the convergence and the expected performance. The choice of the clustering technique remains highly dependent and relative to the modeled task in the SG. In general, it remains unclear what is the best clustering approach/criteria. However, in comparing the clustering based on: (1) load profiles and (2) attributes of the clients for the task of load forecasting, the authors of [47] demonstrated that load profiles provide the best results further confirmed in reference [49].

### 3.4. Security Mechanism

Despite the privacy-preserving character of FL, recent contributions demonstrated that it suffers from several security issues [50, 51]. As the energy data is subject to strict regulations [52], this aspect becomes of paramount importance for the different actors of the power grid. Various security mechanisms were adopted in related work, as illustrated in Figure 3. Authentication is the simplest security mechanism among the analyzed frameworks, implemented commonly through a third-party server [53, 54]. An attribute access control mechanism was suggested in reference [55]. These suggestions introduce a new hop to existing topologies, a trusted third-party server, inducing an extra step where the clients are granted access to avoid malicious injections. The authenticated clients are assumed to be fully trusted.

Despite their complexity, cryptography algorithms were leveraged in several contributions to protect the weights of local models upon transfer through the network. Two main trends of encryption schemes can be identified in related work: (1) encrypt/decrypt scheme and (2) homomorphic encryption. The first involves using a private/public key pair to encrypt the model before exchanging it through the network, implemented with different approaches including a KDC server [56], Paillier encryption [15], and hashing and RSA cryptography [35]. Alternatively, homomorphic encryption allows performing calculations directly on the encrypted model as suggested in reference [57, 58, 59]. While the first alternative is robust against False Data Injection (FDI), the second is more optimized and prevents computational delays. Therefore, adopting one of these two techniques depends highly on the requirements of the service offered. For example, homomorphic encryption would be more suitable for short-term residential load forecasting due to time Constraints.

Differential privacy is another technique adopted to prevent alternating the model weights by introducing a smaller amount of noise to the data and adopting robust aggregation algorithms in the face of potential attacks [60, 61, 62, 63, 64, 55]. For example, the authors of [55] suggest that adopting the differential privacy algorithm can effectively solve the risk of maliciously reconstructing user load data addressed through adversarial networks. Nonetheless, the experimental setup was limited, where a potential enhancement would involve investigating the effect of more robust aggregation algorithms. A main drawback of this technique is its high dependency on the amount of noise added to deliver an acceptable level of security that could negatively impact the performance, as demonstrated in reference [63] for the case of load disaggregation.

Blockchain technology was extensively adopted in the case of EV, with differences in the proposed framework. A common suggestion was integrating the aggregator in a blockchain network [65, 36, 66]. For example, the authors of [65] replace the aggregator with a blockchain network, and the EV fleet is used both as a consumer and a supplier of electrical energy. A primary limitation can be perceived when the number of blocks increases as the system's efficiency will be negatively affected due to huge memory requirements, slow transactions, and mining speed. An alternative would include integrating the EV fleet, and the aggregator in the blockchain network as suggested in reference [67]. In this case, EV clients are deployed to act as miners, which implies the need for more resources on edge devices and that the computing resources on the EV may not be sufficient. A more generic FL-framework for SGs leveraging blockchain technology was suggested in reference [45] to enhance the security of the plethora of energy services. Alliance chain, a variant of blockchain technology, was also adopted in reference [68] as a general FL-a framework for SGs. Other security mechanisms were adopted in related work, such as injection attacks through outlier detection [69] and security control as suggested per [70]. Furthermore, a combination of several previously presented security mechanisms was suggested in reference [59].

Despite their benefits in protecting user's data and the model's weight, all the previously presented security mechanisms add a computational burden on the edge devices, and a compromise between the security level, the available resources, and the performance is undoubtedly required in real SG environments.

## 4. Applications of FL in energy services

Adopting FL for training ML models in energy scholarship recently received significant attention. The conducted literature search revealed five major services, mainly: (1) generic frameworks for SG, (2) load forecasting, (3) Renewable Energy (RE) production, (4) FDI and anomaly detection, (6) Non-Intrusive Load Monitoring (NILM). The application of FL in NILM was already reviewed in reference [23], and thus it is only briefly discussed along with four other services receiving less attention in the last section.

### 4.1. Generic Frameworks

The new SG eco-system gathers ESP and Energy Data Owner (EDO) in a distributed setup. The energy data is thus often stored in isolated islands, which prevents unlocking the full potential of data-driven approaches. The relevance of FL in addressing this issue was highlighted in reference [71, 76], in combination with digital twins enabling real-time simulation and decision-making for the case of [71]. The framework suggests a full exploitation of available sensing technologies with federated learning for real-time power grid management. A similar study was presented in reference [73] investigating the applicability of FL approach in the case of smart buildings to evaluate the convergence time. Moreover, a theoretical contribution suggested in reference [77] recommends an inclusive FL of different grid operators and customers to achieve robust systems against cyber-attacks. Overall, three main research streams for general applications in SG can be identified: (1) performance of the FL framework and its topology, (2) edge resources allocation, and (3) security aspects. Table 2 illustrates a representative set of generic frameworks for SG. It could be observed that all frameworks yield competitive results where the minimum accuracy of 90% and a maximum MAE of 0.03 were recorded. A common limitation, however, was the lack of evaluation of communication overhead and processing resources required on edge devices.

A generic FL framework was proposed in reference [78] for the metering infrastructure suggesting two aggregation schemes: a 2-tiers scheme and a 3-tier scheme reflecting a hierarchical aggregation scheme. The main advantage of the latter is the grouping of the clients based on their geo-localization, validated considering the case of NILM. It revealed equivalent performance to centralized alternatives with significant network load improvements. The impact of the aggregation algorithm was further discussed in reference [79], where an aggregation algorithm considering data characteristics was used to calculate the weight distribution and meet the needs of each node. Another generic implementation of FL in SGs was suggested in reference [15]. The authors investigated both VFL and HFL, tested in the case of predicting consumption patterns of consumers. Two different models were implemented for the vertical framework: a linear model and gradient-boosting decision trees. Nonetheless, the experimental results revealed that the first one is more suitable when a trusted third-party server is required. In contrast, the second one is more suitable when the main focus is on the accuracy of the generated predictions. The authors of [34] formulated an optimization problem to contribute to solving FL tasks (e.g., load forecasting) requested by the ESP. The optimization problem aims at maximizing the profit of both EDO and ESP through two payoff functions. The first one offers incentives from the ESP to convince EDO to participate in the FL task, and the second one represents the gain that the ESP will achieve after paying the incentives. A main limitation of this framework is that it requires the deployment of edge aggregators grouping nearly located EDO and processing delays to address the optimization problem before executing any task, which may consist of an obstacle for real-time processing.

**Table 2**
Representative FL-based contributions for general applications in SGs

| Work | Year | Model | Characteristics | Security | Dataset | Metrics | Results | Main Limitations |
|---|---|---|---|---|---|---|---|---|
| [34] | 2022 | DRL | HFL RL hier. | 2-tier aggregation scheme | Private dataset | Acc. | 0.99 | Potential privacy leakage from the aggregator |
| [71] | 2022 | DNN | HFL S hier. | - | Private dataset | Carbon reduction | 0.49 | Gradient delay due to computational complexity |
| [45] | 2022 | CNN based model | HFL S P2P | Blockchain | Data from Toronto | RMSE MAPE | 0.03 0.03 | Limited evaluation of the energy trading and sharing system |
| [38] | 2022 | RL | HFL RL client-server | | Private dataset | Reward | -10 | Limited evaluation |
| [72] | 2021 | GRU | HFL S client-server | - | Private dataset | Processing delay | 5 | Lack of evaluation on real-data |
| [73] | 2021 | Deep model | HFL S client-server | - | Quarnot's data | MAE | 0.02 | The system can easily be hacked through the reporter |
| [74] | 2021 | ML model | HFL S client-server | Disturbing mechanism | Private dataset | - | - | Limited experimental setup |
| [59] | 2021 | DNN | HFL S | Blockchain | MNIST SVHN | Acc. | 0.98 | 2D transform induces extra computations |
| [75] | 2021 | | HFL client-server | Group signatures | Private dataset | Time | 10.92 | Induces a heavy computational load over the SG |
| [61] | 2021 | ATT-BLSTM | HFL S client-server | DF-privacy | Dataport | MAPE nMAE nRMSE | 0.29 0.04 0.08 | Poor results compared to non-DP schemes |
| [15] | 2021 | XGBoost | VFL HFL S client-server | Paillier-encryption | Data from Zhuhai | MSE | 0.2 | Limited evaluation |
| [68] | 2021 | | HFL S client-server | Alliance chain | MNIST | Acc. | 0.90 | Requires long training time |

The heterogeneity of edge devices in SG, with different computational capabilities, highlights the importance of resource allocation for efficient exploitation. It follows that FL framework, mainly running on these devices, would best consider this constraint for better computational efficiency. This aspect was considered in several related works [71, 72, 80]. The characteristics of the task at hand were considered in [72] as cutting criteria to choose the appropriate offloading strategy. These criteria include the data size and the resources required for computing the task at edge devices leading to either local, hybrid, or distant execution. This framework provides thus optimal exploitation of resources yet may leak privacy with remote and hybrid execution. The previous idea was extended in reference [71] with a dynamic allocation performed at all levels of a hierarchical FL framework. The dynamic nature of run-times was considered in reference [80], leveraging a collaborative adaptive approach to adjust threat detection threshold, adaptive security management, and adaptive models. The suggested method also includes an explainable AI module providing decision support. In the same direction, the authors of [19] consider the problem of communication reliability addressed through timers limiting each step of the FL framework calculation, allowing to overcome link failure and local training delay.

The third research stream (i.e., security) considers different mechanisms, including blockchain, differential privacy, and encryption [81]. Blockchain networks were adopted in both [59, 68]. Another FL framework adopting blockchain was suggested in reference [82] with a distilled knowledge loss overcoming the single point of failure related to the central node. The evaluation has shown good accuracy yet revealed extra communication overhead and sensitivity to intrusion attacks. Alliance chain, a particular case of blockchain, was adopted in reference [68] for safe storage and data usage by power equipment. A key advantage of using an alliance chain is the registration process forcing each power equipment to register before enabling its participation in the learning process. Differential privacy was adopted in reference [60] through a communication protocol to reach a trade-off between resource consumption, user utility, and local differential privacy. This mechanism allows adding noise before uploading the local models, leading to a more robust learning process against intrusion attacks, where the amount depends on the user class ranging from sensitive to regular users. In the same direction, a differential private model aggregation scheme was proposed in reference [61] combined with an attentive model to detect malicious models. With the same goal of securing the FL-SG, the authors of [83] suggest adding an intermediate encryption level between the edge devices and cloud servers. The method reinforces the security of the model parameters by encrypting the process relying on secure fusion methods, including four techniques: (1) differential privacy, (2) secure multiparty computing, (3) homomorphic encryption, and (4) function encryption. Alternatively, group-based signatures were suggested in reference [75] to protect the identity of the grid operators and consumers.

## 4.2. Residential Load Forecasting

Load forecasting is the task of predicting future load demand/generation based on historical load data. It is crucial for efficiently managing power grids [84]. Data-driven techniques were at the core of the load forecasting scholarship [85] in recent years. Centralized approaches raise privacy concerns since load energy data holds sensitive information about the usage of appliances and occupancy data [86]. Furthermore, insufficient data for training is a major obstacle. FL is a viable solution to overcome these issues [87]. Table 3 summarises the main FL-based load forecasting contributions, indicating that horizontal FL is a common design choice for all the existing contributions, with the majority of them following a supervised learning paradigm under a client-server topology. Nonetheless, some exceptions to this trend can be observed (e.g., [39, 88]). These contributions can be grouped according to two main themes: (1) addressing the non-IIDness of the data and (2) enhancing the security of FL frameworks.

The problem of non-IID nature originates from the different behaviors of clients reflecting on their load profiles and leading to convergence issues. It can be addressed using different techniques, such as clustering techniques, personalization of the global model, and client selection strategies. Other techniques include considering an augmented set of features as input to the model (e.g., weekly information[47]) to improve the forecasting accuracy. In this regard, the aggregation techniques were also investigated in reference [92] considering two algorithms, the FedSVG, and the FedAVG. The study concluded that FedAVG, with several training steps of GD on the client's level, provides better forecasting results requiring fewer training rounds to achieve convergence.

To address the data silos problem in power grids, several contributions adopted clustering algorithms. Different features were found to have different effects on the performance. These features include clients attribute [90, 47], the characteristics of the buildings [93], and load profiles [47, 49], with evidence about the superiority of the last approach for load forecasting [47, 49]. Yet, leveraging the building's characteristics [93] revealed better transferability. Global model personalization is another well-acknowledged strategy to address the non-IID nature of data in FL. It was adopted in two different contributions [32, 33] and demonstrated significant enhancement but remains highly dependent on the resources available at the client's level.

Both [95] and [54] propose to select a random set of clients to contribute to the training at each epoch to enable sampling from a more homogeneous data distribution. The experimental setup presented in reference [95] highlighted that using only a subset of clients allows for building powerful forecasting models. A similar framework was presented in [88] leveraging a peer-to-peer topology with a gradient selection strategy. This framework would induce a larger communication overhead than all the previously presented approaches since it would require each client to broadcast its local model. Thus, the authors suggested custom broadcast frequencies for each client, leading to an equivalent performance of centralized learning.

While some scholars [94] argue that design choices such as average pooling already add an extra security layer to standard FL frameworks, they still suffer from several security limitations. For example, an analysis of their sensitivity to poisoning attacks was presented in reference [89], concluding that they are vulnerable. Outlier detection approaches [69] or unsupervised clustering [96] can be used prior to aggregation to detect malicious updates. The encryption of the model's weight was highlighted in [58, 47]. The difference between these contributions is the assumptions made on the trustworthiness of the central server. Assuming that the latter is trusted but curious adds more privacy but can lead to performance loss, as suggested in related work [97]. Little attention was given to this aspect in reference [58]. Several scholars also considered differential privacy [74, 55, 62] where it is used to disturb the weights of the model. In this case, the privacy budget is expected to have minor effects on the overall performance [62], potentially enhanced with attribute-based access control [55]. Despite the security enhancement these mechanisms provide, they induce extra computational and communication resources leading to delays in short-term forecasting. Finding the perfect trade-off between the security level and the required performance remains an open research question.

## 4.3. Renewable Energy

RE sources are crucial in the new power grid for eco-friendly energy production. They represent the best alternative to achieve smart and sustainable energy production. However, wind and PV systems are not controllable regarding their output, which requires careful planning based on accurate prediction of power generation and consumption. Several scholars focused on improving the performance of predicting power generation through several existing techniques. The use of Generative Adversarial Networks (GANs) was suggested in reference [98] combined with the least square error to solve the problem of training instability of these models. The model demonstrated promising results in the case of renewable scenario generation due to its powerful generative modeling ability, where the generated scenarios almost perfectly reproduce the characteristics of the real data while maintaining its diversity. However, a main limitation of this contribution is it assumes all edge devices have the same computational resources. The authors of [48] focus on the learning scheme, suggesting attributing a membership value to each solar power generation station. This membership value is calculated based on the centroids of each cluster, considering the features of the local stations. The suggested FL process thus allows the grouping of similar stations leading to more accurate predictions. A more general FL framework, considering both energy consumers and producers, was suggested in reference [99] to predict power consumption and generation. The framework is beneficial for load swings and load curtailment detection.

**Table**
Rèpresentative FL-based contributions for load forecasting contributions.

| Work | Year | Model | Characteristics | Security | Dataset | Metrics | Results | Main Limitations |
|---|---|---|---|---|---|---|---|---|
| [89] | 2021 | LSTM | HFL S client-server | - | Private data | MAPE | 0.13 | Investigated only two noise additive attacks |
| [32] | 2021 | LSTM | HFL TFL S client-server | - | DATAPORT | RMSE MAPE | 0.49 0.33 | No protection against intrusion attacks |
| [87] | 2021 | LR | HFL S client-server | - | Private dataset | MAPE | 0.11 | Limited experimental setup |
| [49] | 2021 | LSTM | HFL S client-server | - | Australian dataset | RMSE MAPE MAE | 0.12 0.44 0.07 | Potential privacy-leakage |
| [90] | 2021 | LSTM | HFL S client-server | - | Hue dataset | Huber loss | 0.007 | Limited experimental setup |
| [54] | 2021 | DNN | HFL S client-server | - | ASHARAE dataset | RMSE | 1.13 | Assumes that the third party server is trusted |
| [88] | 2021 | LSTM | HFL TFL S P2P | - | DATAPORT | Acc. | 0.99 | Requires heavy computations on edge devices |
| [42] | 2022 | LSTM | HFL S hier. | - | Private data | RMSE | 0.41 | Dependent on edge resources |
| [69] | 2021 | LSTM | HFL S hier. | Outlier detection | DATAPORT | RMSE MAPE | 0.55 0.34 | Suffers from challenges of security robustness and resource optimisation |
| [91] | 2021 | LSTM | HFL S hier. | - | British Data | RMSE | 0.01 | Potential privacy leakage during inference |
| [47] | 2021 | LSTM | HFL S client-server | - | Data from UK | RMSE | 0.12 | Requires hyper-parameters tuning and heavy computations on edge devices. |
| [33] | 2022 | - | HFL TFL S client-server | - | Irish CER dataset | RMSE MAE | 0.82 0.50 | High communication and computation burden |
| [92] | 2022 | LSTM | HFL S client-server | - | Hydro Lodon | RMSE MAPE | 0.61 0.14 | Requires several training epochs on edge devices |
| [93] | 2022 | ANN | HFL S client-server | Encryption of the weights | Data Genome Project | RMSE MAPE | 0.03 0.15 | Not applicable to tree-based algorithms such as decision trees |
| [58] | 2022 | LSTM | HFL S client-server | Encryption of the weights | Data from Ausgrid | MSE | 0.33 | A potential privacy leakage between the edge and the energy retailer |
| [62] | 2022 | LSTM | HFL S client-server | DF-privacy | IEEE Voltage Feeder | QL | 0.032 | No evaluation of potential information leakage |
| [94] | 2022 | CNN | HFL S client-server | Security control mechanism | Data from Tomsk | RMSE MAE MAPE | 0.006 0.002 0.08 | Only evaluated forecasting performance |

Security aspects of the FL framework when considering the prediction of power production are also crucial since the latter includes power trading between different grid operators. To address this aspect, the use of a key distribution center was suggested in reference [56], allowing local clients to encrypt the weights of the model before uploading them to the central server. This framework adds an extra security layer but requires more communication and computational resources at the edge devices. Considering the case of predicting wind power generation, an FDI attack can be modeled through scaling attacks applied to the input data where the wind speed is suggested by [70]. The previous suggestion, tested in different regions, demonstrated an enhanced transferability performance.

**Table 4**
Representative FL-based contributions for RE.

| Title | Year | Model | Features | Learning | Topology | Security | Dataset | Metrics | Main Limitations |
|---|---|---|---|---|---|---|---|---|---|
| [98] | 2022 | least square GANs | HFL | S | client-server | - | Dataset from NREL | MAE 0.17<br>RMSE 0.24 | Assumes optimal communication and the same resources for all edge devices |
| [48] | 2022 | FCN | HFL | S | clustered client-server | - | AMS solar energy | Acc. 0.70 | The clustering was the main concern |
| [99] | 2022 | NN | HFL | S | client-server | - | DataPort | RMSE 1.98 | Lack of comparison with centralised-baseline |
| [56] | 2022 | LSTM | HFL | S | client-server | Cryptography | Readings by Ausgrid | RMSE 0.60<br>MAE 0.33 | No evaluation of the encryption's effect on the overall performance |
| [70] | 2022 | CNN | TFL | S | client-server | Security control | Dataset from IRAN | RMSE 0.02<br>MAE 0.007<br>MAPE 0.64<br>$R^2$ 0.85 | The image processing technique induces heavy computations on edge devices |
| [57] | 2022 | CNN, LSTM | HFL | S | client-server | Homomorphic encryption | Data from SCADA | F1 0.93<br>AUC 0.90<br>MCC 0.50 | Requires heavy computational resources due to the sliding window approach |

Despite the promising results in the previously presented contributions, fault and anomaly detection remains a prominent challenge for adopting renewable energies in distributed power grids [100, 101]. These power sources require continuous monitoring and maintenance to prevent damages [102]. Nonetheless, a major obstacle to deploying fault detection systems is the unbalanced data [103]. This problem becomes more relevant when considering FL as local models being trained with smaller portions of unbalanced data. The previous problem was mitigated in the case of blade icing in wind turbines by balancing the extracted features in the latent space protected through homomorphic encryption [57]. Table 4 summarizes the main FL-based RE frameworks. Despite that these frameworks yield good performance, the table shows that most assume perfect communication conditions, and little attention was given to evaluating other aspects of federated frameworks.

## 4.4. Electrical Vehicles

The adoption of EV is gaining increasing momentum. However, this vehicle's power control and energy management type is still in its infancy, and many problems are yet to be addressed. AI and communication technologies are both key tools for efficient power prediction. Table 5 portrays representative FL-based contributions for electrical vehicle studies. The best-reported results for each contribution reveal a maximum RSME of 6.5 and a minimum accuracy of 90%, which indicates that using FL does not lead to significant deterioration in the performance. However, this remains subject to the considered experimental design (e.g., if the data is IID).

Considering EV demand learning (EDL), three different FL topologies were suggested in reference [104]: (1) a centralized EDL (CEDL) for scenarios where charging stations have limited hardware, (2) federated EDL (FEDL) to protect the data collection process, and (3) clustering-based EDL where CS are grouped into $K$ cluster to minimize the cost of biased predictions with CEDL or FEDL applied on each cluster independently. The latter demonstrated the best results considering both RMSE and communication overhead. The effect of clustering was further investigated in reference [105] based on the similarity of historical energy demand and geo-localization, revealing that the first clustering criteria showed superior performance. An unsupervised clustering based on the client's attribute was suggested in reference [106] through a two-scale regression model. The comparison between all these clustering techniques in the case of EV remains an open research question, and further research is needed in this regard.

Two main concepts emerged to enhance energy trading in EV networks: Vehicle-to-Grid (V2G) and Vehicle-to-Vehicle (V2V). The first concept (i.e., V2G) involves exploiting the power available in EV to feed the power grid. The combination of this concept with FL-based algorithms were first proposed in reference [65] to predict EV power consumption for the next period leading to straight estimation of the power that can be supplied by the EV given the remaining power and the grid's request. This concept was further explored in reference [107] where an adaptive and cost-friendly privacy preservation mechanism for wireless charging FL in V2G systems was proposed with a RL mechanism for local learning. The second concept (i.e., V2V) allows direct energy between FL. It was considered in a two-step learning paradigm in reference [40] to find the optimal charging/discharging variant among V2V, V2G, and G2V through price negotiation. The consideration of V2V revealed an increase of 6,1% in self-sufficiency compared to state-of-the-art methods. The concept of smart contracting was also suggested in reference [108, 109] to solve the problem of competing CS revealing that 24% of enhancement can be achieved in the power demand prediction with a 48% and 36% enhancement in the utilities and social welfare over traditional economic models [108].

VFL was exploited in reference [35, 36] to complete the client's feature set to generate charging recommendations. Both contributions suggest training two models in a synchronized manner where the EV and the CS wait for each other at each epoch before updating the models. An encryption alignment technique was adopted to find the intersection between both local datasets. However, the framework suggested in reference [36] considers the protection of the aggregator inside a blockchain to secure and optimize the calculations. Thus, the latter offers a higher level of security.

The security of FL-based frameworks remains critical and is subject to several possible improvements. The authors of [53] proposed a security-enhanced mutual authentication FL-based energy management framework for EV infrastructure. In the proposed method, cryptographic authentication is applied to prevent unauthorized access. A trust value and award are used to exclude low-prediction or suspect CS and convince CS to contribute to the training. Blockchain is another promising technology that can be leveraged to address some of the issues of FL. Replacing the central node with a blockchain network where a consensus committee is established to update the global model is a viable suggestion, as suggested in reference [66]. Alternatively, [67] suggests integrating the Virtual Power Plant (VPP) aggregator, FL fleets, and a group of miners into the blockchain network. This last contribution also suggests adopting an FL-QLMS algorithm to select a qualified set of models for aggregation, resulting in a more accurate global model. Following this approach, the blockchain network is only adopted after the first aggregation occurs on the VPP level.

### 4.5. False Data Injection and Anomaly Detection

Despite the fact that FL frameworks allow protecting data confidentiality, their security remains an open research topic [51] due to their vulnerability to several attacks originating from the distributed nature of the learning process. False data injection attacks are the most widespread attacks where a client can alternate the global model by updating malicious weights. Table 6 illustrates a representative set of FL-based for FDI detection. The best-recorded performance for these frameworks reveals values of the F1-score between 0.87 and 0.99 highlighting good detection performance for different datasets. However, most of these frameworks suppose the availability of annotated data on edge devices which can be hard to obtain in real setups.

A demonstration of such an attack in SG setups was suggested in reference [114], also leveraging side channel attack to build a dataset similar to the benign clients. The study resulted in countermeasures serving as recommendations for future implementations. These recommendations include: (1) the necessity to protect physical access to the hardware, (2) Adding randomization to the model, and (3) the necessity to use FDI detection techniques. The third countermeasure was central to recent research contributions [115, 116].

An FL-based approach for intrusion detection in the metering infrastructure was suggested in reference [115, 116] where the conducted studies demonstrated acceptable results. Both contributions were evaluated on the same dataset and resulted in equivalent results. This aspect was further explored for Photovoltaic(s) (PV) power prediction [37, 114], leveraging techniques such as customization and asynchronous learning to enhance the overall system. Alternatively, the authors of [117] propose a custom aggregation scheme based on the geometric median and the Weisfeld's algorithm [118] allowing to mitigate the effect of noisy, irregular gradient and reducing the overall communication overhead. The latter was also a main concern in reference [119]. However, the authors addressed this problem through a compressed system log reducing the transmission delay and increasing privacy protection.

Other malicious attacks in the case of the power grid may include abnormal behavior detection. For this purpose, an FL-based approach with LSTM models was used in reference [120] in smart buildings. This framework follows a multi-task learning paradigm leading to the reduction of the processing delay. Energy theft, a particular case of abnormal behavior, was investigated in reference [64], where each detection station gathers data from different households, and the learning process is further protected with differential privacy during local training. The evaluation setup of this method demonstrated that it had a very low communication overhead while outperforming state-of-the-art models even with differential privacy.

### 4.6. Other Applications

FL application is not restricted in the aforementioned use cases. Other use cases have also adopted this learning paradigm but with less attention. These energy services include: load disaggregation, thermal comfort control, pattern identification, energy recommendation, demand response, and smart energy trading and contracting.

Energy disaggregation (or NILM) is the set of approaches aiming to identify individual loads related to operating devices using as measured by a smart meter [122, 123]. NILM scholarship received significant attention in the past decade marked by several turning points. Adopting deep models to solve the problem of load disaggregation in 2015 [124] was a major breakpoint. Typically, a tremendous number of contributions followed [125]. The common practice in these approaches is the centralized training of the models, where it is assumed that the data is gathered at a central point. Nonetheless, such practice can violate a consumer's privacy as information about his daily routines [126, 127] will be exposed. FL was adopted in few contributions [128, 129, 130, 131, 132, 133, 63] to address this issue. Most of these contributions reveal good results with IID data but reported a significant decrease in the performance with non-IID data. The only contribution revealing promising results was presented in reference [133], where a comparison between locally-trained, centrally-trained, and federated models was performed.

**Table**
Representative FL-based contributions for EV and CS

| Title | Year | Model | Features | Learning | Topology | Security | Dataset | Metrics | Results | Main Limitations |
|---|---|---|---|---|---|---|---|---|---|---|
| [104] | 2019 | DL | HFL TFL | S | client-server | - | Dundee, UK | RMSE | 5.87 | Clustering only considered location |
| [65] | 2020 | MLP | HFL | S | client-server | Blockchain | Data from Japan | $R^2$ | 0.92 | The efficiency of the system decreases with an increasing number of blocks |
| [110] | 2020 | S | HFL | - | client-server | - | - | MLE* | 3 | Assumes IID data |
| [79] | 2021 | CNN | HFL | S | client-server | Trusted third-party | MNIST dataset | Acc. | 0.97 | Considers a small number of nodes |
| [53] | 2021 | | HFL | | hier. | Third-party server | CIFAR-10 and MNIST | Acc. | 0.90 | Provides poor results with few number of adversarial clients |
| [35] | 2021 | | VFL | S | P2P | Hashing and RSA | Private dataset | AUC | 0.94 | Considered an evaluation dataset without the user's behavior |
| [106] | 2021 | NN | HFL | S | client-server | - | Private dataset | MCRPS<br>RMSE | 0.56<br>0.98 | No personalisation technique considered for individual drivers |
| [111] | 2021 | LSTM | HFL | S | client-server | - | private dataset | MAE | | Limited set of features were considered during validation |
| [108] | 2022 | DNN | HFL | S | clustered client-server | - | Dundee, UK | RMSE | 6.5 | Clustering only considered location |
| [112] | 2022 | RL | HFL | S | client-server | - | Private dataset | Reward | -2.05 | Enhancement only observed on the long term |
| [67] | 2022 | | HFL | S | client-server | Blockchain | Private dataset | RMSE | 5.2 | Dependent on computational resources of the edge devices |
| [113] | 2022 | CNN LSTM | HFL | S | client-server | - | DATAPORT<br>REDD<br>REFIT<br>UKDALE | Acc.<br>Rec.<br>Prec.<br>F1 | 0.97<br>0.93<br>0.75<br>0.81 | Requires extra computational resources on edge devices |

Applying FL in thermal comfort control received attention from the authors of [134]. They suggested using FL in two different applications related to smart buildings, namely thermal comfort modeling and short-term forecasting. The experiments proved the efficiency of FL, showing a high prediction accuracy. Moreover, the authors assessed the effect of the personalization step, demonstrating that it evolves the trained model enough to fit all participants. Thermal comfort prediction also received interest from [135]. Using FL solves both over-fitting and privacy-revealing problems. The FedAvg algorithm is used after being endowed with a branch selection protocol.

As providing customized services to householders is the ultimate objective of retailers, identifying household electricity consumption patterns becomes crucial. However, retailers' privacy represents a major concern when dealing with problems requiring data sharing. The authors of [136] suggested using FL for consumer profile identification in a privacy-preserving way. Considering this framework, the training stage is accelerated by utilizing an asynchronous stochastic gradient descent with delay compensation (ASGD-DC) to achieve the global model update. ASGD-DC helps optimize the global parameters update and facilitate the model's performance. The privacy-preserving nature further motivated the use of FL in [137] to propose an electricity consumer profile identification method through a three-step classification algorithm. The evaluation phase showed that using PCA-based feature extraction enhanced the identification model performance indicating that the federated model has comparable performance whether trained on the balanced or unbalanced dataset.

Energy recommendation is another application where FL bears great potential. The study presented in reference [5] discusses the synergy between big data and FL to analyze collected data and provide timed recommendations to the user. The project's ultimate objective is to change user habits by leveraging the concept of micro-moments. The FL showed a prediction accuracy comparable to the centralized learning models. Additionally, it outperforms those models in privacy-preserving and time performance, revealing that FL scales up to large environments.

**Table**
Representative FL-based contributions for FDI detection in SG

| Title | Year | Model | Features | Learning | Topology | Security | Dataset | Metrics | Results | Main Limitations |
|---|---|---|---|---|---|---|---|---|---|---|
| [119] | 2022 | LSTM | HFL | S | client-server | - | Private dataset | Acc. | 0.93 | Potential privacy leakage on edge servers |
| [115] | 2022 | FCN | HFL | S | client-server | - | NSL-KDD | Acc. | 0.99 | Suffers from security issues |
| [116] | 2021 | DNN | HFL | S | client-server | | NSL-KDD | Prec.<br>Rec.<br>F1 | 0.99<br>0.98<br>0.99 | Provides merely acceptable results in the case of classes with few samples |
| [121] | 2022 | LSTM | HFL | S | client-server | - | Hourly electricity load, Toronto | Acc. | 0.89 | Suppose physical access to capture power trace |
| [120] | 2021 | LSTM | HFL | S | client-server | - | Sensors Event Log dataset | Prec.<br>Rec.<br>F1 | 0.89<br>0.79<br>0.87 | Suffers from security issues |
| [114] | 2022 | CNN | HFL | S | P2P | - | Private dataset | Acc. | 0.99 | Convergence issues and communication overhead with increasing number of clients |
| [37] | 2022 | LSTM-based model | HFL | S | client-server | - | Private dataset | Acc.<br>Prec.<br>Rec.<br>F1 | 0.97<br>0.95<br>0.95<br>0.96 | Only considered the case of PVs |
| [64] | 2022 | TCN-based model | HFL | S | client-server | DF-privacy | SGCC dataset | Acc. | 0.92 | The third party server induce a delay in the computations |
| [117] | 2022 | GAN | HFL | SS | client-server | Robust aggregation scheme | Simulated dataset | Acc.<br>F1<br>AUC | 0.97<br>0.97<br>0.99 | Requires heavy computations on the edge devices |

Demand response is a mechanism to balance energy production and consumption by customers reducing their electricity usage during times of peak demand. As indicated in [138], demand response can be addressed using FL. In particular, demand response requires customers changing their energy consumption behavior to lower critical-peak demand by opting for off-peak energy consumption or changing the energy source [139, 140]. In this regard, [39] focused on regulating electricity production and demand. The FL is used for distributed learning to tackle the privacy-revealing problem. The authors adopted a deep RL algorithm to handle uncertainty issues. Furthermore, non-convex power flow constraints were handled by transforming the updating neural network parameters into a sequence of semi-definite programs. The proposed algorithm reduces both peak load and the user's daily cost. Moreover, the algorithm convergence was evaluated as good as it converges to the centralized algorithm solution.

Energy sharing enables energy exchange between consumer and prosumer communities in return for future benefits [45]. In [45], the authors suggest an autonomous smart contracting system for SG leveraging the blockchain-based FL scheme to estimate the power demand of the consumers and power produced by prosumers. The P2P sharing of the energy allows for efficient exploitation of RE in microgrids. In the same context of energy sharing, [141] discussed the challenges facing prosumer communities preventing them from achieving collective goals. For this purpose, the authors developed a framework based on FL due to its distributed nature leading to better collaboration within the prosumer community. The main objective of this work is to balance the usage of different energy sources (PV and V2G). FL guarantees privacy protection and avoids over-fitting in the energy production/consumption prediction process. Likewise, the collaboration within the prosumer community was considered in [142] aiming at optimizing energy consumption in large-scale grids. The authors suggested exploiting FL to allow agents to collaboratively find optimization solutions. A hybrid (synchronous/asynchronous) global parameter update is used to reduce communication costs. More precisely, agents are grouped into clusters where the update inter-cluster is asynchronous while the update intra-cluster is synchronous.

**Table 7**
Challenges facing FL in SGs

| Challenge | Aspect of FL | Source | Possible solutions |
|---|---|---|---|
| **Limited and heterogeneous edge-devices** | Topology | Heterogeneity of hardware used by different grid operators | - Smart selection of clients<br>- Model compression techniques<br>- Adaptive resource allocation |
| **Training Bottleneck and convergence time** | ML training | - Non-identically distributed data<br>- Communication overhead | - Clustering of clients<br>- Robust aggregation algorithms<br>- Use of timers |
| **Inference Attacks** | Security & privacy | Inference of training data used from the FL updates | Secure computations (e.g., homomorphic encryption)/ Differential privacy / Third party server |
| **Poisoning Attacks** | Security & privacy | Data poisoning / model poisoning | FDI detection before aggregation / Blockchain |
| **Backdoor Attacks** | Security & privacy | Introduction of backdoor by a set of devices to target the labeling of a single task | Differential privacy / Norm thresholding updates |
| **Evaluation protocols** | Overall FL scheme | No standardized FL evaluation framework | Standardization of evaluation protocols/ Development of new evaluation protocols for FL in SGs |

## 5. Open Challenges

Despite the extensive research efforts to adapt different variants of FL to the requirements of SG, adopting this learning paradigm in real setups remains subject to several challenges. Different technical issues still await additional investigations and research, such as privacy and security concerns and incentive mechanism design. In this section, we present some challenges and promising research directions to overcome these problems, improve the performance of FL models and widen their adoption in SG. Table 7 summarizes these challenges.

The first challenge is the heterogeneity of edge devices and their limited hardware in some scenarios. This challenge opposes the main goal of SG to deliver near real-time services. Particularly, when training tasks are initiated on devices with different computational capabilities, a delay could occur since the aggregation process cannot start before receiving the updated weights from all the clients. With this in mind, potential solutions could consist in using offloading strategies with a smart selection of the clients based on their hardware characteristics. Furthermore, compression techniques can be leveraged to address the problem of computational time. Resource management is thus a main issue in distributed learning. In addition, this problem imposes compatibility issues, so a standardization effort is required. Using advanced strategies for the dynamic allocation of resources can also introduce a considerable computational delay at the beginning of each training task. Furthermore, failures could occur during the training and induce the failure of the whole process. Further research on this topic is thus required with appropriate evaluation protocols that consider all the previously mentioned scenarios to deliver a clearer idea about the robustness of FL in the case of SG.

The training bottleneck and convergence time are also main challenges for FL in SG, mainly caused by the heterogeneity of data in different devices or the distributed nature of the learning process. More precisely, it is widely acknowledged that different energy clients exhibit different consumption patterns directly reflecting on their energy curve and thus hindering learning convergence, especially with the high number of local training epochs. One possible solution to avoid this issue is clustering clients (i.e., HFL). Nonetheless, there is little evidence in the reviewed literature about the best clustering criteria for the case of different energy applications. Only a few studies provided a comparative evaluation of different clustering criteria for the possibility of load forecasting. Yet, the evaluation was limited since the comparison only considered the performance and the convergence time disregarding other aspects. For example, the model's updates transmitted at every training epoch can divulge clients' private data. In this regard, an adversary can infer the training data through the weights updates transferred at each iteration. Moreover, multiple malicious attacks may threaten the models learned by the servers. They can poison the models or delay/prevent their convergence.

Simultaneously ensuring security protection and privacy preservation in FL while maintaining high models' accuracy and low computational cost is a critical issue [143]. Specifically, while ensuring rigorous privacy assurances is of utmost importance, mechanisms deployed to preserve clients' data privacy need to moderately hold the accuracy rates of the learned models. Along the same line, privacy-preserving techniques must not significantly increase the computational overhead of the training phase or introduce an excessive overhead to the network [144, 145]. In addition, while protection strategies against poisoning and Byzantine attacks need the FL server to analyze individual client-supplied updates [146], privacy-enhancing methodologies using secure Multiparty Computation (MPC) involve significant overhead and conceal the individual updates from the FL servers [147]. This is because they usually rely on heavy encryption to combine local updates before employing them in the global model. Thus, they preclude the servers from measuring weight statistics and accuracy metrics on individual updates. Hence detecting and discarding malicious updates can not be reached [148]. It is unarguable that including all the clients in each training round is not feasible in real scenarios as it would overwhelm the communication infrastructure. Thus, FL approaches generally rely on the selection strategies of different clients. Several selection strategies that exist were previously described.

# 6. Future Directions

Different FL technical issues, such as privacy and security concerns and incentive mechanism design, still await additional investigations and research efforts. Developing techniques for FL on non-IID data is one of the critical research directions. FL algorithms typically assume that the data at each location is IID. However, in many real-world settings, data may be non-IID due to factors such as location or demographics. Research is needed to develop FL techniques that handle non-IID data.

To overcome these challenges, the following future directions can be considered:

## 6.1. Model Pruning

Model pruning can reduce the size and complexity of DL models, making them more suitable for edge devices. In FL, participating devices often have limited resources, making it challenging to train large models. By pruning the model, researchers can remove redundant parameters, giving the model more freedom to adapt to the different calculation and communication capabilities of the clients in the FL process [149, 150].

## 6.2. Secure multiparty computation

It can allow multiple parties to jointly compute the result of a function using their private inputs without revealing these inputs to one another. In the context of FL, the secure MPC can be used to protect the privacy of user updates, but it can also increase communication volume and computational complexity. Finding a balance between privacy protection and communication costs is a vital field research area [151]. Furthermore, MPC is a cryptographic technique that enables distributed computing systems to securely aggregate local models that contain sensitive information. While MPC can prevent direct leak computation-efficient, the server may still be able to indirectly obtain local information in some cases or even reveal the true images through methods such as Deep Leakage from Gradients (DLG) [152].

## 6.3. Asynchronous online FL

Although FedAvg is the most popular method for optimizing FL, it assumes unrealistic device homogeneity. To overcome this issue, asynchronous online FL, a technique that addresses the challenges of data and equipment heterogeneity in FL on distributed edge devices, can be used [153]. This approach addresses issues such as device load, lag, or withdrawal, and some studies have proposed using active device selection to address device heterogeneity. Moreover, while standard FL assumes that clients have offline access to their data samples generated statistically, the study suggested in reference [154] breaks away from this assumption and explores FL in uncertain environments, where clients' local loss functions arrive in an online streaming manner without statistical assumptions. The study uses a collective regret metric as the performance measure, departing from the traditional FL approach where clients have offline access to their data samples generated statistically. Another proposed method for handling incomplete local updates involves scaling aggregation coefficients, but the effectiveness of this approach has not yet been demonstrated. Alternatively, [155] discusses minimizing the training latency of a wireless FL system while maintaining client data privacy. Hence, a client scheduling scheme is used to reduce the number of training rounds and time intervals by jointly considering the significance of each client's local updates and delay issues. The problem is formulated as a multi-armed bandit program, and an online scheduling scheme based on the $\epsilon$-greedy algorithm is proposed to achieve a tradeoff between exploration and exploitation.

## 6.4. Interpretability and explainability

Interpretability and explainability are other important features in FL-based SG frameworks. Typically, explainable VFL improves the performance and security of AI systems that deal with low-quality data [156, 157]. This can include a credibility assessment strategy, a federated counterfactual explanation, and an importance rate metric [158]. In this regard, FedeX, an FL-based anomaly detection solution for edge-based industrial control systems, was introduced in reference [159]. It uses XAI for interpretability, allowing experts to make quick decisions and trust the model more. [160] proposes a lifecycle dashboard as a solution to address the need for explainability in FL-based systems by considering the requirements and perspectives of SG stakeholders, visualizing information from the FL server and being generic enough to be applied to different use cases and industries. In summary, the potential impact of FL is ultimately improving the SG by (i) helping break through the inherent information exchange barriers and (ii) allowing for all the SG parties to trustingly collaborate on energy data mining, with an enhanced level of privacy protection is very promising.

# 7. Conclusion

This study contributes to the literature with a detailed overview of FL applications in SG considering different energy services. To the best of our knowledge, this work is the first to analyze the development of FL in SG through a detailed analysis of design trends considering different aspects with a discussion of representative contributions for popular energy services. Furthermore, the review highlighted the need for a holistic approach to SG management and the potential of FL to facilitate efficient and secure collaboration among different actors in the energy sector. Only few technical questions have been answered, and FL is expected to be an active research area throughout the next decade.

## References

[1] M. Y. Mehmood, A. Oad, M. Abrar, H. M. Munir, S. F. Hasan, H. Muqeet, N. A. Golilarz, Edge computing for IoT-enabled smart grid, Security and Communication Networks 2021 (2021).

[2] A. Alsalemi, Y. Himeur, F. Bensaali, A. Amira, An innovative edge-based internet of energy solution for promoting energy saving in buildings, Sustainable Cities and Society 78 (2022) 103571.

[3] A. Sayed, Y. Himeur, A. Alsalemi, F. Bensaali, A. Amira, Intelligent edge-based recommender system for internet of energy applications, IEEE Systems Journal 16 (3) (2021) 5001–5010.

[4] Y. Himeur, A. Alsalemi, F. Bensaali, A. Amira, I. Varlamis, G. Bravos, C. Sardianos, G. Dimitrakopoulos, Techno-economic assessment of building energy efficiency systems using behavioral change: A case study of an edge-based micro-moments solution, Journal of Cleaner Production 331 (2022) 129786.

[5] I. Varlamis, C. Sardianos, C. Chronis, G. Dimitrakopoulos, Y. Himeur, A. Alsalemi, F. Bensaali, A. Amira, Using big data and federated learning for generating energy efficiency recommendations, International Journal of Data Science and Analytics (2022) 1–17.

[6] A. N. Sayed, F. Bensaali, Y. Himeur, M. Houchati, Edge-based real-time occupancy detection system through a non-intrusive sensing system, Energies 16 (5) (2023) 2388.

[7] Y. Himeur, A. Alsalemi, F. Bensaali, A. Amira, The emergence of hybrid edge-cloud computing for energy efficiency in buildings, in: Intelligent Systems and Applications: Proceedings of the 2021 Intelligent Systems Conference (IntelliSys) Volume 2, Springer, 2022, pp. 70–83.

[8] J. Konečný, H. B. McMahan, F. X. Yu, P. Richtárik, A. T. Suresh, D. Bacon, Federated learning: Strategies for improving communication efficiency, arXiv preprint arXiv:1610.05492 (2016).

[9] J. Ogier du Terrail, A. Leopold, C. Joly, C. Béguier, M. Andreux, C. Maussion, B. Schmauch, E. W. Tramel, E. Bendjebbar, M. Zaslavskiy, et al., Federated learning for predicting histological response to neoadjuvant chemotherapy in triple-negative breast cancer, Nature Medicine (2023) 1–12.

[10] L. Ouyang, F.-Y. Wang, Y. Tian, X. Jia, H. Qi, G. Wang, Artificial identification: a novel privacy framework for federated learning based on blockchain, IEEE Transactions on Computational Social Systems (2023).

[11] S. Singh, S. Rathore, O. Alfarraj, A. Tolba, B. Yoon, A framework for privacy-preservation of IoT healthcare data using federated learning and blockchain technology, Future Generation Computer Systems 129 (2022) 380–388.

[12] M. A. Ferrag, O. Friha, D. Hamouda, L. Maglaras, H. Janicke, Edge-IIoTset: A new comprehensive realistic cyber security dataset of IoT and IIoT applications for centralized and federated learning, IEEE Access 10 (2022) 40281–40306.

[13] Y. Shen, F. Gou, J. Wu, Node screening method based on federated learning with iot in opportunistic social networks, Mathematics 10 (10) (2022) 1669.

[14] L. U. Khan, W. Saad, Z. Han, E. Hossain, C. S. Hong, Federated learning for internet of things: Recent advances, taxonomy, and open challenges, IEEE Communications Surveys & Tutorials (2021).

[15] H. Liu, X. Zhang, X. Shen, H. Sun, A federated learning framework for smart grids: Securing power traces in collaborative learning, arXiv preprint arXiv:2103.11870 (2021).

[16] M. L. Tuballa, M. L. Abundo, A review of the development of smart grid technologies, Renewable and Sustainable Energy Reviews 59 (2016) 710–725.

[17] W. Y. B. Lim, N. C. Luong, D. T. Hoang, Y. Jiao, Y.-C. Liang, Q. Yang, D. Niyato, C. Miao, Federated learning in mobile edge networks: A comprehensive survey, IEEE Communications Surveys & Tutorials 22 (3) (2020) 2031–2063.

[18] X. Yin, Y. Zhu, J. Hu, A comprehensive survey of privacy-preserving federated learning: A taxonomy, review, and future directions, ACM Computing Surveys (CSUR) 54 (6) (2021) 1–36.

[19] S. Zhai, X. Jin, L. Wei, H. Luo, M. Cao, Dynamic federated learning for GMEC with time-varying wireless link, IEEE Access 9 (2021) 10400–10412.

[20] A. Li, J. Sun, X. Zeng, M. Zhang, H. Li, Y. Chen, FedMask: Joint computation and communication-efficient personalized federated learning via heterogeneous masking, in: Proceedings of the 19th ACM Conference on Embedded Networked Sensor Systems, SenSys ’21, Association for Computing Machinery, 2021, pp. 42–55, event-place: Coimbra, Portugal.

[21] A. Nilsson, S. Smith, G. Ulm, E. Gustavsson, M. Jirstrand, A performance evaluation of federated learning algorithms, in: Proceedings of the second workshop on distributed infrastructures for deep learning, 2018, pp. 1–8.

[22] D. Chai, L. Wang, K. Chen, Q. Yang, FEDEVAL: A benchmark system with a comprehensive evaluation model for federated learning, arXiv preprint arXiv:2011.09655 (2020).

[23] H. Bousbiat, Y. Himeur, I. Varlamis, F. Bensaali, A. Amira, Neural load disaggregation: Meta-analysis, federated learning and beyond, Energies 16 (2) (2023).

[24] Q. Yang, Y. Liu, T. Chen, Y. Tong, Federated machine learning: Concept and applications, ACM Transactions on Intelligent Systems and Technology (TIST) 10 (2) (2019) 1–19.

[25] P. M. Mammen, Federated learning: opportunities and challenges, arXiv preprint arXiv:2101.05428 (2021).

[26] D. C. Nguyen, M. Ding, P. N. Pathirana, A. Seneviratne, J. Li, H. V. Poor, Federated learning for internet of things: A comprehensive survey, IEEE Communications Surveys & Tutorials 23 (3) (2021) 1622–1658.

[27] L. Li, Y. Fan, M. Tse, K.-Y. Lin, A review of applications in federated learning, Computers & Industrial Engineering 149 (2020) 106854.

[28] Z. Sun, W. Hu, C. Li, Cross-lingual entity alignment via joint attribute-preserving embedding, in: International Semantic Web Conference, Springer, 2017, pp. 628–644.

[29] Z. Chen, W. Liao, P. Tian, Q. Wang, W. Yu, A fairness-aware peer-to-peer decentralized learning framework with heterogeneous devices, Future Internet 14 (5) (2022) 138.

[30] Z. Yu, J. Hu, G. Min, H. Xu, J. Mills, Proactive content caching for internet-of-vehicles based on peer-to-peer federated learning, in: 2020 IEEE 26th International Conference on Parallel and Distributed Systems (ICPADS), IEEE, 2020, pp. 601–608.

[31] I. Hegedűs, G. Danner, M. Jelasity, Gossip learning as a decentralized alternative to federated learning, in: IFIP International Conference on Distributed Applications and Interoperable Systems, Springer, 2019, pp. 74–90.

[32] Y. Xu, C. Jiang, Z. Zheng, B. Yang, N. Zhu, LSTM short-term residential load forecasting based on federated learning, in: 2021 International Conference on Mechanical, Aerospace and Automotive Engineering, CMAAE 2021, Association for Computing Machinery, 2021, pp. 217–221, event-place: Changsha, China.

[33] Y. Wang, N. Gao, G. Hug, Personalized federated learning for individual consumer load forecasting, CSEE Journal of Power and Energy Systems (2022).

[34] Z. Su, Y. Wang, T. H. Luan, N. Zhang, F. Li, T. Chen, H. Cao, Secure and efficient federated learning for smart grid with edge-cloud collaboration, IEEE Transactions on Industrial Informatics

18 (2) (2022) 1333–1344.
[35] X. Wang, X. Zheng, X. Liang, Charging station recommendation for electric vehicle based on federated learning, in: Journal of physics: Conference series, Vol. 1792, IOP Publishing, 2021, p. 012055.
[36] Z. Teimoori, A. Yassine, M. S. Hossain, A secure cloudlet-based charging station recommendation for electric vehicles empowered by federated learning, IEEE Transactions on Industrial Informatics 18 (9) (2022) 6464–6473.
[37] L. Zhao, J. Li, Q. Li, F. Li, A federated learning framework for detecting false data injection attacks in solar farms, IEEE Transactions on Power Electronics 37 (3) (2021) 2496–2501.
[38] H. Liu, W. Wu, Federated reinforcement learning for decentralized voltage control in distribution networks, IEEE Transactions on Smart Grid 13 (5) (2022) 3840–3843.
[39] S. Bahrami, Y. C. Chen, V. W. S. Wong, Deep reinforcement learning for demand response in distribution networks, IEEE Transactions on Smart Grid 12 (2) (2021) 1496–1506.
[40] Y. Tao, J. Qiu, S. Lai, X. Sun, Y. Wang, J. Zhao, Data-driven matching protocol for vehicle-to-vehicle energy management considering privacy preservation, IEEE Transactions on Transportation Electrification (2022) 1–1.
[41] Y. Wang, M. Jia, N. Gao, L. Von Krannichfeldt, M. Sun, G. Hug, Federated clustering for electricity consumption pattern extraction, IEEE Transactions on Smart Grid 13 (3) (2022) 2425–2439.
[42] N. Gholizadeh, P. Musilek, Federated learning with hyperparameter-based clustering for electrical load forecasting, Internet of Things 17 (2022) 100470.
[43] J. C. Jiang, B. Kantarci, S. Oktug, T. Soyata, Federated learning in smart city sensing: Challenges and opportunities, Sensors 20 (21) (2020) 6230.
[44] Y. He, F. Luo, G. Ranzi, W. Kong, Short-term residential load forecasting based on federated learning and load clustering, in: 2021 IEEE International Conference on Communications, Control, and Computing Technologies for Smart Grids (SmartGridComm), IEEE, 2021, pp. 77–82.
[45] O. Bouachir, M. Aloqaily, O. Ozkasap, F. Ali, FederatedGrids: Federated learning and blockchain-assisted p2p energy sharing, IEEE Transactions on Green Communications and Networking 6 (1) (2022) 424–436.
[46] Y. L. Tun, K. Thar, C. M. Thwal, C. S. Hong, Federated learning based energy demand prediction with clustered aggregation, in: 2021 IEEE International Conference on Big Data and Smart Computing (BigComp), IEEE, 2021, pp. 164–167.
[47] M. Savi, F. Olivadese, Short-term energy consumption forecasting at the edge: A federated learning approach, IEEE Access 9 (2021) 95949–95969.
[48] E. Yoo, H. Ko, S. Pack, Fuzzy clustered federated learning algorithm for solar power generation forecasting, IEEE Transactions on Emerging Topics in Computing (2022).
[49] Y. He, F. Luo, G. Ranzi, W. Kong, Short-term residential load forecasting based on federated learning and load clustering, in: 2021 IEEE International Conference on Communications, Control, and Computing Technologies for Smart Grids, SmartGridComm 2021, 2021, pp. 77–82.
[50] Y. Himeur, S. S. Sohail, F. Bensaali, A. Amira, M. Alazab, Latest trends of security and privacy in recommender systems: a comprehensive review and future perspectives, Computers & Security (2022) 102746.
[51] V. Mothukuri, R. M. Parizi, S. Pouriyeh, Y. Huang, A. Dehghantanha, G. Srivastava, A survey on security and privacy of federated learning, Future Generation Computer Systems 115 (2021) 619–640.
[52] M. R. Asghar, G. Dán, D. Miorandi, I. Chlamtac, Smart meter data privacy: A survey, IEEE Communications Surveys & Tutorials 19 (4) (2017) 2820–2835.
[53] W. Wang, M. H. Fida, Z. Lian, Z. Yin, Q.-V. Pham, T. R. Gadekallu, K. Dev, C. Su, Secure-enhanced federated learning for ai-empowered electric vehicle energy prediction, IEEE Consumer Electronics Magazine (2021) 1–1.
[54] S. V. Dasari, K. Mittal, S. GVK, J. Bapat, D. Das, Privacy enhanced energy prediction in smart building using federated learning, in: 2021 IEEE International IOT, Electronics and Mechatronics Conference (IEMTRONICS), 2021, pp. 1–6.
[55] M. Sun, J. Li, Y. Ren, S. Fang, j. Yan, Research on federated learning and its security issues for load forecasting, in: 2021 The 13th International Conference on Computer Modeling and Simulation, ICCMS '21, Association for Computing Machinery, 2021, pp. 237–243, event-place: Melbourne, VIC, Australia.
[56] M. M. Badr, M. I. Ibrahem, M. Mahmoud, W. Alasmary, M. M. Fouda, K. H. Almotairi, Z. M. Fadlullah, Privacy-preserving federated-learning-based net-energy forecasting, in: SoutheastCon 2022, 2022, pp. 133–139.
[57] X. Cheng, F. Shi, Y. Liu, X. Liu, L. Huang, et al., Wind turbine blade icing detection: a federated learning approach, Energy 254 (PC) (2022).
[58] M. A. Husnoo, A. Anwar, N. Hosseinzadeh, S. N. Islam, A. N. Mahmood, R. Doss, FedREP: Towards horizontal federated load forecasting for retail energy providers, arXiv preprint arXiv:2203.00219 (2022).
[59] P. Singh, M. Masud, M. S. Hossain, A. Kaur, G. Muhammad, A. Ghoneim, Privacy-preserving serverless computing using federated learning for smart grids, IEEE Transactions on Industrial Informatics 18 (11) (2022) 7843–7852.
[60] H. Cao, S. Liu, R. Zhao, X. Xiong, Ifed: A novel federated learning framework for local differential privacy in power internet of things, International Journal of Distributed Sensor Networks 16 (5) (2020) 1550147720919698.
[61] X.-Y. Zhang, J. Cordoba-Pachon, C. Watkins, S. Kuenzel, Differential private federated learning for privacy-preserving third party service framework in advanced metering infrastructure, Preprint (2021).
[62] J.-F. Toubeau, F. Teng, T. Morstyn, L. V. Krannichfeldt, Y. Wang, Privacy-preserving probabilistic voltage forecasting in local energy communities, IEEE Transactions on Smart Grid (2022) 1–1.
[63] H. Pötter, S. Lee, D. Mossé, Towards privacy-preserving framework for non-intrusive load monitoring, in: Proceedings of the Twelfth ACM International Conference on Future Energy Systems, e-Energy '21, Association for Computing Machinery, 2021, pp. 259–263, event-place: Virtual Event, Italy.
[64] M. Wen, R. Xie, K. Lu, L. Wang, K. Zhang, FedDetect: A novel privacy-preserving federated learning framework for energy theft detection in smart grid, IEEE Internet of Things Journal 9 (8) (2022) 6069–6080.
[65] Z. Wang, M. Ogbodo, H. Huang, C. Qiu, M. Hisada, A. B. Abdallah, AEBIS: AI-enabled blockchain-based electric vehicle integration system for power management in smart grid platform, IEEE Access 8 (2020) 226409–226421.
[66] B. Li, Y. Guo, Q. Du, Z. Zhu, X. Li, R. Lu, $\mathcal{P}^3$: Privacy-preserving prediction of real-time energy demands in ev charging networks, IEEE Transactions on Industrial Informatics (2022).
[67] Z. Wang, A. Ben Abdallah, A robust multi-stage power consumption prediction method in a semi-decentralized network of electric vehicles, IEEE Access 10 (2022) 37082–37096.
[68] P. Fang, C. Sun, Y. Lian, Z. Wang, F. Xiao, L. Chen, P. Lin, Data security sharing mechanism of power equipment based on federated learning, in: Q. Liu, X. Liu, B. Chen, Y. Zhang, J. Peng (Eds.), Proceedings of the 11th International Conference on Computer Engineering and Networks, Springer Nature Singapore, Singapore, 2022, pp. 875–883.
[69] R. Firouzi, R. Rahmani, T. Kanter, Federated learning for distributed reasoning on edge computing, in: Procedia Computer Science, Vol. 184, 2021, pp. 419–427.
[70] H. Moayyed, A. Moradzadeh, B. Mohammadi-Ivatloo, A. Aguiar, R. Ghorbani, A cyber-secure generalized supermodel for wind power forecasting based on deep federated learning and image processing,

Energy Conversion and Management 267 (2022).
[71] H. Liao, Z. Zhou, N. Liu, Y. Zhang, G. Xu, Z. Wang, S. Mumtaz, Cloud-edge-device collaborative reliable and communication-efficient digital twin for low-carbon electrical equipment management, IEEE Transactions on Industrial Informatics 19 (2) (2023) 1715–1724.
[72] Y. Zhu, J. Xu, Y. Xie, J. Jiang, X. Yang, Z. Li, Dynamic task offloading in power grid internet of things: A fast-convergent federated learning approach, in: 2021 IEEE 6th International Conference on Computer and Communication Systems (ICCCS), 2021, pp. 933–937.
[73] A. Mitra, Y. Ngoko, D. Trystram, Impact of federated learning on smart buildings, in: 2021 International Conference on Artificial Intelligence and Smart Systems (ICAIS), IEEE, 2021, pp. 93–99.
[74] G. Si, Y. Zhang, Y. Sun, Privacy protection strategy based on federated learning for smart park multi energy fusion system, in: 2021 IEEE 4th International Conference on Computer and Communication Engineering Technology, CCET 2021, 2021, pp. 393–395.
[75] S. Kanchan, A. Kumar, A. S. Saqib, B. J. Choi, Group signature based federated learning in smart grids, in: 2021 7th International Conference on Signal Processing and Intelligent Systems (ICSPIS), 2021, pp. 1–5.
[76] Q. Lu, H. Jiang, S. Chen, Y. Gu, T. Gao, J. Zhang, Applications of digital twin system in a smart city system with multi-energy, in: 2021 IEEE 1st International Conference on Digital Twins and Parallel Intelligence (DTPI), 2021, pp. 58–61.
[77] S. You, A cyber-secure framework for power grids based on federated learning, Preprint (2020).
[78] N. Hudson, M. J. Hossain, M. Hosseinzadeh, H. Khamfroush, M. Rahnamay-Naeini, N. Ghani, A framework for edge intelligent smart distribution grids via federated learning, in: 2021 International Conference on Computer Communications and Networks (ICCCN), 2021, pp. 1–9.
[79] M. Wang, Y. Lei, Y. Liang, X. Lv, W. Mo, Agi-fedavg: A uav power line inspection algorithm based on federated learning, in: 2021 IEEE 3rd International Conference on Civil Aviation Safety and Information Technology (ICCASIT), 2021, pp. 1030–1034.
[80] S. Boudko, H. Abie, E. Nigussie, R. Savola, Towards federated learning-based collaborative adaptive cybersecurity for multi-microgrids., in: WINSYS, 2021, pp. 83–90.
[81] Y. Himeur, A. Alsalemi, F. Bensaali, A. Amira, A. Al-Kababji, Recent trends of smart nonintrusive load monitoring in buildings: A review, open challenges, and future directions, International Journal of Intelligent Systems 37 (10) (2022) 7124–7179.
[82] G. Huang, C. Wu, Y. Hu, C. Guo, Serverless distributed learning for smart grid analytics, Chinese Physics B 30 (8) (2021) 088802.
[83] H. Yang, Y. Bai, Z. Zou, Y. Shi, R. Yang, Research on the security sharing model of power grid data based on federated learning, in: 2021 IEEE 5th Information Technology, Networking, Electronic and Automation Control Conference (ITNEC), Vol. 5, IEEE, 2021, pp. 1566–1569.
[84] H. Hou, C. Liu, Q. Wang, X. Wu, J. Tang, Y. Shi, C. Xie, Review of load forecasting based on artificial intelligence methodologies, models, and challenges, Electric Power Systems Research 210 (2022) 108067.
[85] J. Zhu, H. Dong, W. Zheng, S. Li, Y. Huang, L. Xi, Review and prospect of data-driven techniques for load forecasting in integrated energy systems, Applied Energy 321 (2022) 119269.
[86] K. Panchabikesan, F. Haghighat, M. El Mankibi, Data driven occupancy information for energy simulation and energy use assessment in residential buildings, Energy 218 (2021) 119539.
[87] S. Zhang, Z. Xu, J. Wang, J. Chen, Y. Xia, Improving the accuracy of load forecasting for campus buildings based on federated learning, in: 2021 IEEE International Conference on Networking, Sensing and Control (ICNSC), Vol. 1, 2021, pp. 1–5.
[88] J. Gao, W. Wang, Z. Liu, M. F. R. M. Billah, B. Campbell, Decentralized federated learning framework for the neighborhood: A case study on residential building load forecasting, in: Proceedings of the 19th ACM Conference on Embedded Networked Sensor Systems, SenSys '21, Association for Computing Machinery, 2021, pp. 453–459, event-place: Coimbra, Portugal.
[89] N. B. S. Qureshi, D.-H. Kim, J. Lee, E.-K. Lee, On the performance impact of poisoning attacks on load forecasting in federated learning, in: Adjunct Proceedings of the 2021 ACM International Joint Conference on Pervasive and Ubiquitous Computing and Proceedings of the 2021 ACM International Symposium on Wearable Computers, UbiComp '21, Association for Computing Machinery, 2021, pp. 64–66, event-place: Virtual, USA.
[90] Y. Tun, K. Thar, C. Thwal, C. Hong, Federated learning based energy demand prediction with clustered aggregation, in: Proceedings - 2021 IEEE International Conference on Big Data and Smart Computing, BigComp 2021, 2021, pp. 164–167.
[91] C. Briggs, Z. Fan, P. Andras, Federated learning for short-term residential load forecasting, IEEE Open Access Journal of Power and Energy 9 (2022) 573–583.
[92] M. N. Fekri, K. Grolinger, S. Mir, Distributed load forecasting using smart meter data: Federated learning with recurrent neural networks, International Journal of Electrical Power & Energy Systems 137 (2022) 107669.
[93] J. Li, C. Zhang, Y. Zhao, W. Qiu, Q. Chen, X. Zhang, Federated learning-based short-term building energy consumption prediction method for solving the data silos problem, Building Simulation 15 (6) (2022) 1145–1159.
[94] A. Moradzadeh, H. Moayyed, B. Mohammadi-Ivatloo, A. Aguiar, A. Anvari-Moghaddam, A secure federated deep learning-based approach for heating load demand forecasting in building environment, IEEE Access 10 (2022) 5037–5050.
[95] A. Taïk, S. Cherkaoui, Electrical load forecasting using edge computing and federated learning, in: ICC 2020 - 2020 IEEE International Conference on Communications (ICC), 2020, pp. 1–6.
[96] N. B. S. Qureshi, D.-H. Kim, J. Lee, E.-K. Lee, Poisoning attacks against federated learning in load forecasting of smart energy, in: NOMS 2022-2022 IEEE/IFIP Network Operations and Management Symposium, 2022, pp. 1–7.
[97] H. Fang, Q. Qian, Privacy preserving machine learning with homomorphic encryption and federated learning, Future Internet 13 (4) (2021) 94.
[98] Y. Li, J. Li, Y. Wang, Privacy-preserving spatiotemporal scenario generation of renewable energies: A federated deep generative learning approach, IEEE Transactions on Industrial Informatics 18 (4) (2022) 2310–2320.
[99] V. Venkataramanan, S. Kaza, A. Annaswamy, DER forecast using privacy preserving federated learning, IEEE Internet of Things Journal (2022).
[100] Y. Himeur, K. Ghanem, A. Alsalemi, F. Bensaali, A. Amira, Artificial intelligence based anomaly detection of energy consumption in buildings: A review, current trends and new perspectives, Applied Energy 287 (2021) 116601.
[101] Y. Himeur, A. Alsalemi, F. Bensaali, A. Amira, A novel approach for detecting anomalous energy consumption based on micro-moments and deep neural networks, Cognitive Computation 12 (2020) 1381–1401.
[102] A. Mellit, G. M. Tina, S. A. Kalogirou, Fault detection and diagnosis methods for photovoltaic systems: A review, Renewable and Sustainable Energy Reviews 91 (2018) 1–17.
[103] Y. Himeur, A. Alsalemi, F. Bensaali, A. Amira, Smart power consumption abnormality detection in buildings using micromoments and improved k-nearest neighbors, International Journal of Intelligent Systems 36 (6) (2021) 2865–2894.
[104] Y. M. Saputra, D. T. Hoang, D. N. Nguyen, E. Dutkiewicz, M. D. Mueck, S. Srikanteswara, Energy demand prediction with federated learning for electric vehicle networks, in: 2019 IEEE Global Communications Conference (GLOBECOM), 2019, pp. 1–6.
[105] D. Perry, N. Wang, S.-S. Ho, Energy demand prediction with optimized clustering-based federated learning, in: 2021 IEEE Global Communications Conference (GLOBECOM), IEEE, 2021, pp. 1–6.

[106] A. T. Thorgeirsson, S. Scheubner, S. Fünfgeld, F. Gauterin, Probabilistic prediction of energy demand and driving range for electric vehicles with federated learning, IEEE Open Journal of Vehicular Technology 2 (2021) 151–161.

[107] S. R. Pokhrel, M. B. Hossain, Data privacy of wireless charging vehicle to grid (v2g) networks with federated learning, IEEE Transactions on Vehicular Technology 71 (8) (2022) 9032–9037.

[108] Y. M. Saputra, D. N. Nguyen, D. T. Hoang, T. X. Vu, E. Dutkiewicz, S. Chatzinotas, Federated learning meets contract theory: Economic-efficiency framework for electric vehicle networks, IEEE Transactions on Mobile Computing 21 (8) (2022) 2803–2817.

[109] S. Lee, D.-H. Choi, Dynamic pricing and energy management for profit maximization in multiple smart electric vehicle charging stations: A privacy-preserving deep reinforcement learning approach, Applied Energy 304 (2021) 117754.

[110] S. Samarakoon, M. Bennis, W. Saad, M. Debbah, Distributed federated learning for ultra-reliable low-latency vehicular communications, IEEE Transactions on Communications 68 (2) (2020) 1146–1159.

[111] M. Liu, Fed-bev: A federated learning framework for modelling energy consumption of battery electric vehicles, in: 2021 IEEE 94th Vehicular Technology Conference (VTC2021-Fall), 2021, pp. 1–7.

[112] Q. V. Do, Q.-V. Pham, W.-J. Hwang, Deep reinforcement learning for energy-efficient federated learning in uav-enabled wireless powered networks, IEEE Communications Letters 26 (1) (2021) 99–103.

[113] X. Wang, G. Tang, Y. Wang, S. Keshav, Y. Zhang, Evsense: A robust and scalable approach to non-intrusive ev charging detection, in: Proceedings of the Thirteenth ACM International Conference on Future Energy Systems, e-Energy ’22, Association for Computing Machinery, New York, NY, USA, 2022, p. 307–319.

[114] Q. Liu, B. Yang, Z. Wang, D. Zhu, X. Wang, K. Ma, X. Guan, Asynchronous decentralized federated learning for collaborative fault diagnosis of pv stations, IEEE Transactions on Network Science and Engineering (2022).

[115] H. Liang, D. Liu, X. Zeng, C. Ye, An intrusion detection method for advanced metering infrastructure based on federated learning, Journal of Modern Power Systems and Clean Energy (2022) 1–11.

[116] P. H. Mirzaee, M. Shojafar, Z. Pooranian, P. Asefy, H. Cruickshank, R. Tafazolli, FIDS: A federated intrusion detection system for 5g smart metering network, in: 2021 17th International Conference on Mobility, Sensing and Networking (MSN), 2021, pp. 215–222.

[117] M. Abdel-Basset, N. Moustafa, H. Hawash, Privacy-preserved generative network for trustworthy anomaly detection in smart grids: A federated semi-supervised approach, IEEE Transactions on Industrial Informatics (2022).

[118] F. Plastria, The weiszfeld algorithm: proof, amendments, and extensions, in: Foundations of location analysis, Springer, 2011, pp. 357–389.

[119] S. Hou, J. Lu, E. Zhu, H. Zhang, A. Ye, A federated learning-based fault detection algorithm for power terminals, Mathematical Problems in Engineering 2022 (2022).

[120] R. A. Sater, A. B. Hamza, A federated learning approach to anomaly detection in smart buildings, ACM Transactions on Internet of Things 2 (4) (2021) 1–23.

[121] L. Ding, J. Wu, C. Li, A. Jolfaei, X. Zheng, Sca-lfd: Side-channel analysis based load forecasting disturbance in the energy internet, IEEE Transactions on Industrial Electronics (2022).

[122] H. Bousbiat, C. Klemenjak, W. Elmenreich, Exploring time series imaging for load disaggregation, in: Proceedings of the 7th ACM International Conference on Systems for Energy-Efficient Buildings, Cities, and Transportation, 2020, pp. 254–257.

[123] A. Faustine, L. Pereira, H. Bousbiat, S. Kulkarni, Unet-nilm: A deep neural network for multi-tasks appliances state detection and power estimation in nilm, in: Proceedings of the 5th International Workshop on Non-Intrusive Load Monitoring, 2020, pp. 84–88.

[124] J. Kelly, W. Knottenbelt, Neural NILM: Deep neural networks applied to energy disaggregation, in: Proceedings of the 2nd ACM international conference on embedded systems for energy-efficient built environments, 2015, pp. 55–64.

[125] P. Huber, A. Calatroni, A. Rumsch, A. Paice, Review on deep neural networks applied to low-frequency NILM, Energies 14 (9) (2021) 2390.

[126] H. Bousbiat, C. Klemenjak, G. Leitner, W. Elmenreich, Augmenting an assisted living lab with non-intrusive load monitoring, in: 2020 IEEE International Instrumentation and Measurement Technology Conference (I2MTC), IEEE, 2020, pp. 1–5.

[127] H. Bousbiat, G. Leitner, W. Elmenreich, Ageing safely in the digital era: A new unobtrusive activity monitoring framework leveraging on daily interactions with hand-operated appliances, Sensors 22 (4) (2022) 1322.

[128] Y. Zhang, G. Tang, Q. Huang, Y. Wang, K. Wu, K. Yu, X. Shao, FedNILM: Applying federated learning to NILM applications at the edge, IEEE Transactions on Green Communications and Networking (2022).

[129] R. Schwermer, J. Buchberger, R. Mayer, H.-A. Jacobsen, Federated office plug-load identification for building management systems, in: Proceedings of the Thirteenth ACM International Conference on Future Energy Systems, e-Energy ’22, Association for Computing Machinery, 2022, pp. 114–126, event-place: Virtual Event.

[130] X. Chang, W. Li, A. Y. Zomaya, Fed-GBM: A cost-effective federated gradient boosting tree for non-intrusive load monitoring, in: Proceedings of the Thirteenth ACM International Conference on Future Energy Systems, e-Energy ’22, Association for Computing Machinery, 2022, pp. 63–75, event-place: Virtual Event.

[131] Y. Shi, W. Li, X. Chang, A. Y. Zomaya, User privacy leakages from federated learning in NILM applications, in: Proceedings of the 8th ACM International Conference on Systems for Energy-Efficient Buildings, Cities, and Transportation, BuildSys ’21, Association for Computing Machinery, 2021, pp. 212–213, event-place: Coimbra, Portugal.

[132] Q. Li, J. Ye, W. Song, Z. Tse, Energy disaggregation with federated and transfer learning, in: 7th IEEE World Forum on Internet of Things, WF-IoT 2021, 2021, pp. 698–703.

[133] H. Wang, C. Si, J. Zhao, A federated learning framework for non-intrusive load monitoring, arXiv preprint arXiv:2104.01618 (2021).

[134] M. Khalil, M. Esseghir, L. Merghem–Boulahia, Federated learning for energy-efficient thermal comfort control service in smart buildings, in: 2021 IEEE Global Communications Conference (GLOBECOM), 2021, pp. 01–06.

[135] M. Khalil, M. Esseghir, L. Merghem-Boulahia, A federated learning approach for thermal comfort management, Advanced Engineering Informatics 52 (2022) 101526.

[136] J. Lin, J. Ma, J. Zhu, Privacy-preserving household characteristic identification with federated learning method, IEEE Transactions on Smart Grid 13 (2) (2022) 1088–1099.

[137] Y. Wang, I. L. Bennani, X. Liu, M. Sun, Y. Zhou, Electricity consumer characteristics identification: A federated learning approach, IEEE Transactions on Smart Grid 12 (4) (2021) 3637–3647.

[138] X. Cheng, C. Li, X. Liu, A review of federated learning in energy systems, 2022 IEEE/IAS Industrial and Commercial Power System Asia (I&CPS Asia) (2022) 2089–2095.

[139] W. Huang, N. Zhang, C. Kang, M. Li, M. Huo, From demand response to integrated demand response: Review and prospect of research and application, Protection and Control of Modern Power Systems 4 (1) (2019) 1–13.

[140] A. Sobe, W. Elmenreich, Smart microgrids: Overview and outlook, in: Proceedings of the ITG INFORMATIK Workshop on Smart Grids, Braunschweig, Germany, 2012.

[141] A. Taik, B. Nour, S. Cherkaoui, Empowering prosumer communities in smart grid with wireless communications and federated edge learning, IEEE Wireless Communications 28 (6) (2021) 26–33.

[142] J. Mohammadi, J. Thornburg, Connecting distributed pockets of energy flexibility through federated computations: Limitations and possibilities, in: Proceedings - 19th IEEE International Conference on Machine Learning and Applications, ICMLA 2020, 2020, pp. 1161–1166.

[143] N. Truong, K. Sun, S. Wang, F. Guitton, Y. Guo, Privacy preservation in federated learning: An insightful survey from the gdpr perspective, Computers & Security 110 (2021) 102402.
[144] A. Blanco-Justicia, J. Domingo-Ferrer, S. Martínez, D. Sánchez, A. Flanagan, K. E. Tan, Achieving security and privacy in federated learning systems: Survey, research challenges and future directions, Engineering Applications of Artificial Intelligence 106 (2021) 104468.
[145] M. Asad, A. Moustafa, C. Yu, A critical evaluation of privacy and security threats in federated learning, Sensors 20 (24) (2020) 7182.
[146] D. Cao, S. Chang, Z. Lin, G. Liu, D. Sun, Understanding distributed poisoning attack in federated learning, in: 2019 IEEE 25th International Conference on Parallel and Distributed Systems (ICPADS), IEEE, 2019, pp. 233–239.
[147] P. Liu, X. Xu, W. Wang, Threats, attacks and defenses to federated learning: issues, taxonomy and perspectives, Cybersecurity 5 (1) (2022) 1–19.
[148] N. Bouacida, P. Mohapatra, Vulnerabilities in federated learning, IEEE Access 9 (2021) 63229–63249.
[149] Y. Jiang, S. Wang, V. Valls, B. J. Ko, W.-H. Lee, K. K. Leung, L. Tassiulas, Model pruning enables efficient federated learning on edge devices, IEEE Transactions on Neural Networks and Learning Systems (2022).
[150] P. Prakash, J. Ding, R. Chen, X. Qin, M. Shu, Q. Cui, Y. Guo, M. Pan, Iot device friendly and communication-efficient federated learning via joint model pruning and quantization, IEEE Internet of Things Journal 9 (15) (2022) 13638–13650.
[151] C. Zhang, S. Ekanut, L. Zhen, Z. Li, Augmented multi-party computation against gradient leakage in federated learning, IEEE Transactions on Big Data (2022).
[152] R. Kanagavelu, Q. Wei, Z. Li, H. Zhang, J. Samsudin, Y. Yang, R. S. M. Goh, S. Wang, Ce-fed: Communication efficient multi-party computation enabled federated learning, Array 15 (2022) 100207.
[153] Y. Chen, Y. Ning, M. Slawski, H. Rangwala, Asynchronous online federated learning for edge devices with non-iid data, in: 2020 IEEE International Conference on Big Data (Big Data), IEEE, 2020, pp. 15–24.
[154] A. Mitra, H. Hassani, G. J. Pappas, Online federated learning, in: 2021 60th IEEE Conference on Decision and Control (CDC), 2021, pp. 4083–4090.
[155] B. Xu, W. Xia, J. Zhang, T. Q. Quek, H. Zhu, Online client scheduling for fast federated learning, IEEE Wireless Communications Letters 10 (7) (2021) 1434–1438.
[156] C. Sardianos, I. Varlamis, C. Chronis, G. Dimitrakopoulos, A. Alsalemi, Y. Himeur, F. Bensaali, A. Amira, The emergence of explainability of intelligent systems: Delivering explainable and personalized recommendations for energy efficiency, International Journal of Intelligent Systems 36 (2) (2021) 656–680.
[157] Y. Himeur, M. Elnour, F. Fadli, N. Meskin, I. Petri, Y. Rezgui, F. Bensaali, A. Amira, Ai-big data analytics for building automation and management systems: a survey, actual challenges and future perspectives, Artificial Intelligence Review (2022) 1–93.
[158] P. Chen, X. Du, Z. Lu, J. Wu, P. C. Hung, Evfl: An explainable vertical federated learning for data-oriented artificial intelligence systems, Journal of Systems Architecture 126 (2022) 102474.
[159] T. T. Huong, T. P. Bac, K. N. Ha, N. V. Hoang, N. X. Hoang, N. T. Hung, K. P. Tran, Federated learning-based explainable anomaly detection for industrial control systems, IEEE Access 10 (2022) 53854–53872.
[160] M. Ungersböck, T. Hiessl, D. Schall, F. Michahelles, Explainable federated learning: A lifecycle dashboard for industrial settings, IEEE Pervasive Computing (2023).